\documentclass{article}

\usepackage{PRIMEarxiv}

\usepackage[utf8]{inputenc} 
\usepackage[T1]{fontenc}    
\usepackage{hyperref}       
\usepackage{url}            
\usepackage{booktabs}       
\usepackage{amsfonts}       
\usepackage{nicefrac}       
\usepackage{microtype}      
\usepackage{lipsum}
\usepackage{graphicx}
\graphicspath{{media/}}     
\usepackage{amsmath} 
\usepackage{amssymb} 

\usepackage{rotating} 
\usepackage{threeparttable}
\usepackage{multirow}
\usepackage{tabularx}
\usepackage{makecell}
\usepackage{arydshln}
\usepackage[
  backend=biber,
  style=numeric,   
  sorting=none     
]{biblatex}
\title{Constraint Breeds Generalization: Temporal Dynamics as an Inductive Bias
}

\author{
  Xia Chen \\
  Georg Nemetschek Institute \\
  Munich Data Science Institute \\
  Technische Universität München \\
  \texttt{x.c.chen@tum.de} \\
}

\begin{document}
\maketitle
\setlength{\textfloatsep}{10pt plus 1.0pt minus 2.0pt}
\begin{refsection}

\begin{abstract}

Modern deep learning prioritizes unconstrained scaling and optimization, yet biological systems exploit temporal dynamics under strict metabolic constraints. We propose that dissipative dynamics function as a unique class of inductive bias, distinct from spatial regularizers; temporal phase-space contraction compels networks to discard transient noise and consolidate invariant representations over time. Through phase-space analysis, we identify a critical regime that compels signal evolution to align with neural networks' spectral bias, maximizing generalization across three levels: representational (cross-encoding classification), structural (spontaneous emergence of receptive fields), and behavioral (zero-shot transfer in reinforcement learning). Importantly, this capability requires temporal integration to decode invariants that static architectures perceive as noise. Our findings reframe temporal irreversibility from physical limitation to computational principle: robust AI development requires not merely scaling, but mastering the temporal characteristics that naturally promote generalization.

\end{abstract}


\section{Introduction}

Modern deep learning paradigms largely abstract away the temporal dimension, flattening time into a static spatial index to facilitate efficient parallelization~\cite{krizhevsky2012imagenet, lecun2015deep, hooker2021hardware}. This simplified view contrasts sharply with biological systems, where temporal dynamics are intrinsic to computation under severe energy budget~\cite{buzsaki2006rhythms, shankar2025bridging}. Unlike current AI paradigms that equate generalization with unconstrained scaling~\cite{neyshabur2017exploring, kaplan2020scaling}, the brain's strict metabolic demands force neural dynamics into a dissipative regime characterized by sparse, temporally precise activities~\cite{olshausen1996emergence, attwell2001energy, van2001rate}. Such energy-efficiency selection pressure serves not merely as a limitation, but as a fundamental organizing principle: by naturally compressing the solution space over time, this physical constraint compels the system to abstract robust representations from noisy sensory streams.

This observation motivates our central hypothesis: temporal structure, when properly constrained, functions as an active regularizer for generalization. Specifically, we propose that networks processing temporally encoded signals under dissipative dynamics will learn representations inherently invariant to task-irrelevant variations. To incorporate this principle computationally, we leverage Spiking Neural Networks (SNNs) as the computational substrate, as they provide the necessary theoretical foundation and biological plausibility~\cite{roy2019towards, eshraghian2023training}. Through a sequence of experiments driven by emergent observations, we establish that structured dissipative constraints actively induce stable, invariant features that remain consistently robust across:
\begin{enumerate}
    \item \textit{Representational level}: In cross-encoding classification, networks trained under contractive constraints exhibit asymmetric generalization capability while networks trained under expansive dynamics fail (Experiment 1).
    \item \textit{Structural level}: In unsupervised learning, proper contractive dynamics spontaneously induce biologically plausible, structured receptive fields, whereas other regimes yield isotropic noise (Experiment 2).
    \item \textit{Behavioral level}: In reinforcement learning, where inputs are inherently temporal, we validate the principle through two complementary pathways: encoding-level constraints (consistent with Experiments 1 and 2) and architecture-level constraints (via network internal state dissipation). Both pathways achieve superior zero-shot transfer to unseen physical environments, outperforming baseline architectures (Experiment 3). 
\end{enumerate}

Beyond these empirical findings, we discover that this generalization capability requires a co-dependence of signal dynamics and networks capable of temporal integration: conventional time-indexing architectures fail to benefit from this principle even when provided with identical inputs. Mechanistically, spectral analysis reveals that the proper contractive signal trajectory (termed "transition regime") emerges as a "low-frequency, high-entropy" signature, aligning with the spectral bias of neural networks~\cite{rahaman2019spectral}. These results suggest that dynamical constraints act as a distinct class of \textit{inductive bias}~\cite{goyal2022inductive} that exploits temporal irreversibility as a critical computational principle. Unlike explicit, hand-crafted spatial regularizers (e.g., data augmentation~\cite{hernandez2018data, balestriero2022effects} or architectural priors~\cite{battaglia2018relational}), this implicit dissipative process operates through physical phase-space contraction (defined by $\Sigma\lambda_i$), suggesting that robust generalization may emerge not from removing constraints, but from computationally mastering them.

\section{Experiments}
\subsection{Method: Dynamical Constraints via Dual Pathways}

To test whether this principle is implementation-agnostic, we realize dissipative constraints through two complementary pathways that both govern the same physical quantity: the global Lyapunov sum ($\Sigma\lambda_i$), which quantifies phase space contraction during signal evolution~\cite{wolf1985determining}. Prior work has established that downstream computational capability depends primarily on this global contraction rate rather than local chaotic divergence ($\lambda_{max}$)~\cite{drgovna2022dissipative, chen2025dynamical}.

\textit{Encoding-Level Constraints.}
For tasks with static inputs (Experiments 1-2), we transform features $\mathbf{x} \in \mathbb{R}^d$ into temporal trajectories using a parametrically controllable duffing oscillator system~\cite{kovacic2011duffing}. This design allows us to tune the global phase space dynamics via a single parameter $\delta$ within the same system to avoid geometric confounds inherent when comparing distinct chaotic systems. Each input dimension initializes a three-dimensional dynamical system that evolves for time $T$ with $N$ discrete steps, producing trajectories $\phi_\delta(\mathbf{x}, t) \in \mathbb{R}^{d \times N \times 3}$. The parameter $\delta$ controls the system's phase space behavior:
\begin{itemize}
    \item \textit{Expansive} ($\delta < 0$): Divergent dynamics ($\Sigma\lambda_i > 0$) that amplify initial conditions.
    \item \textit{Transition} ($\delta = 0-2.0$): Weakly dissipative dynamics ($\Sigma\lambda_i \lesssim 0$). 
    \item \textit{Dissipative} ($\delta \gg 0$):  Contractive dynamics ($\Sigma\lambda_i \ll 0$) that suppress state evolution.
\end{itemize} 

The trajectory's origin is invariant to $\delta$; the parameter solely governs the subsequent temporal evolution of the signal. Therefore, tuning $\delta$ enables us to explicitly modulate the degree of constraint imposed on the temporal structure (see Appendix \ref{app:exp} for detailed dynamical justification).

\textit{Architecture-Level Constraints.}
For tasks with inherently temporal inputs (Experiment 3), we leverage the SNN's intrinsic dissipation of Leaky Integrate-and-Fire (LIF) neurons~\cite{eshraghian2023training}. The membrane potential evolves as:$$\text{mem}_{t+1} = \beta \cdot \text{mem}_t + \text{input}_t$$ where the leak parameter $\beta$ controls the effective integration window ($\tau_{\text{mem}} = -1/\ln(\beta)$), acting as the architectural counterpart to the encoding parameter $\delta$: high $\beta$ ($\approx 1.0$) preserves information across timesteps (weak constraint), while low $\beta$ ($\approx 0.1$) rapidly dissipates internal states (strong constraint). This setup aims to validate that our findings derive from the constraint principle itself, rather than properties specific to input encoding.

\subsection{Experiment 1: Generalization in Cross-Encoding Classification}
\label{sec:generalization}

We first investigate \textit{representational transfer}~\cite{yosinski2014transferable, deng2022strong} as a direct proxy for learned invariance. The rationale is straightforward: if dissipative constraints lead to the abstraction of robust representations, a network trained on this specific dynamical regime should generalize to others, analogous to domain adaptation but operating across the spectrum of temporal dynamics.

Specifically, we designed a cross-encoding evaluation protocol using the handwritten digits dataset (64 features, 10 classes)~\cite{optical_recognition_of_handwritten_digits_80}. SNNs with LIF neurons are benchmarked against recurrent architectures and static baselines: Long Short-Term Memory networks (LSTMs)~\cite{hochreiter1997long}, Recurrent Neural Networks (RNNs), Multi-Layer Perceptrons (MLPs) processing only the final timestep (\textit{Last-T}), and MLPs processing time-averaged inputs (\textit{Avg-Pool}). All networks share an identical three-layer, fully-connected topology to isolate algorithmic differences. We sampled 12 dynamical regimes from $\delta \in [-1.5, 10.0]$ and constructed a full $12 \times 12$ generalization matrix, training each architecture on one regime ($\delta_{\text{train}}$) and evaluating across all others ($\delta_{\text{test}}$). Detailed experiment settings are provided in Appendix~\ref{app:exp1}.

\begin{figure}[ht!]
    \centering

    \includegraphics[width=0.92\textwidth]{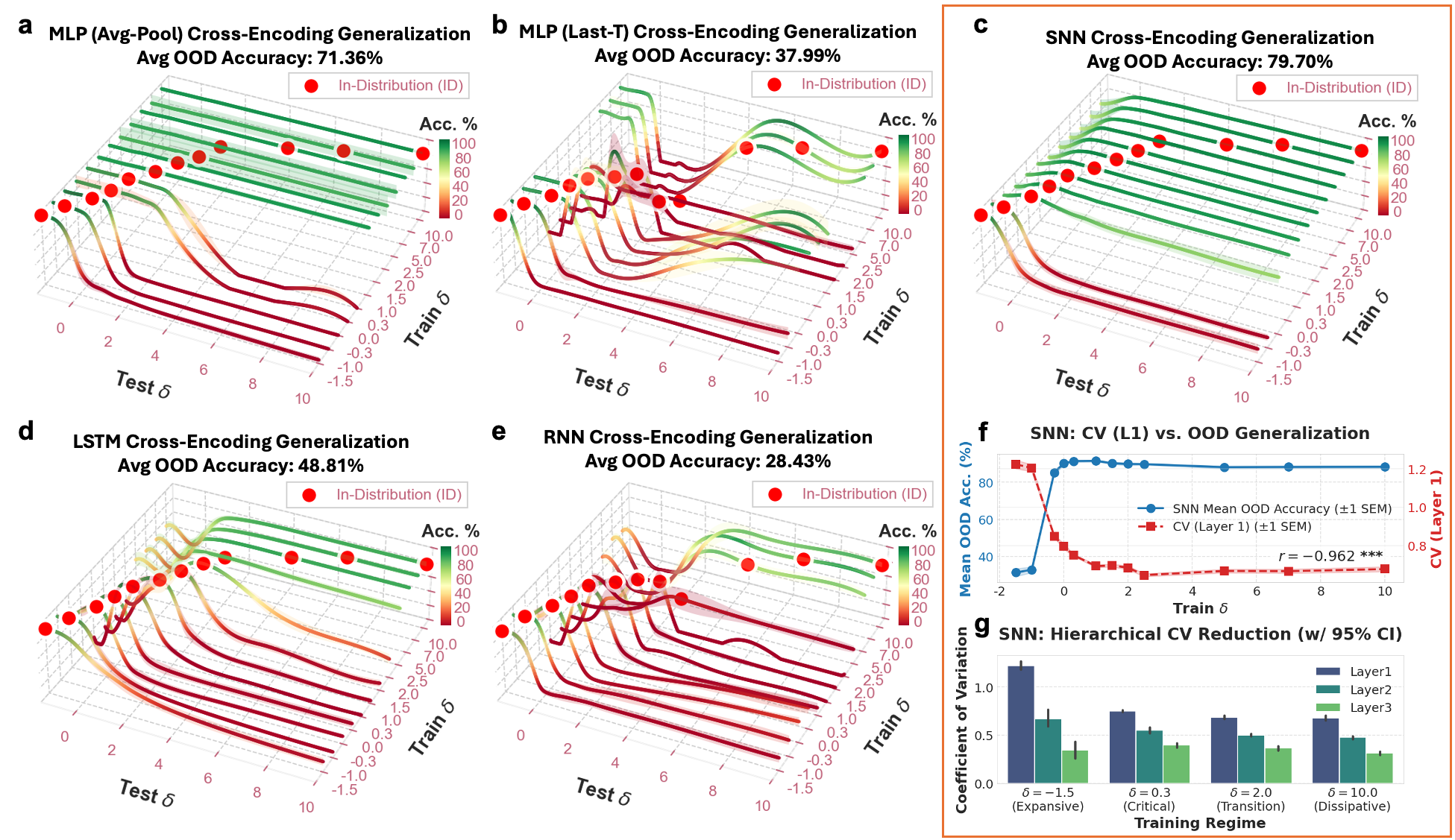} 
    \caption{
      \textbf{Cross-encoding generalization with hierarchical variance reduction.} 
      \textbf{(a-e)} 3D generalization landscapes (mean accuracy with deviation, $n=10$). 
      \textbf{(c)} SNNs exhibit a robust generalization in the transition regime ($\delta \in [0, 2]$) across the spectrum. This contrasts with the diagonal-only ridges seen in expansive regimes ($\delta < 0$) and in baseline architectures (\textit{Last-T MLP, LSTM, RNN}; \textbf{b, d, e}), and \textit{Avg-Pool MLP} (\textbf{a}) shows partial, unstable generalization (see Appendix Tables \ref{tab:cross_enc_mlp_avg}--\ref{tab:cross_enc_rnn}). 
      \textbf{(f)} Out-of-Distribution accuracy strongly anti-correlates with Layer 1 neural variability (CV; $r=-0.962$). Error bars: $\pm 1$ SEM. 
      \textbf{(g)} Hierarchical variance reduction (L1 $\to$ L3) emerges in SNN across regimes.
    }
    \label{fig:1}
\end{figure}

\textbf{Asymmetric Generalization Landscapes.} 
Figure~\ref{fig:1} presents generalization performance as a 3D topographic landscape. We define In-Distribution (ID) accuracy as the performance on the training encoding (the diagonal ridges where $\delta_{\text{train}} \approx \delta_{\text{test}}$), and Out-of-Distribution (OOD) accuracy as the performance across the remaining spectrum. 
While all architectures maintain high ID performance (ridge peaks), SNNs (Figure~\ref{fig:1}c) reveal a unique asymmetry: Networks trained in expansive regimes ($\delta_{\text{train}} < 0$) form sharp, narrow ridges that achieve high ID accuracy but collapse immediately when $\delta_{\text{test}} > 0$. This behavior is characteristic of memorization without transfer. In contrast, SNNs trained in the transition ($\delta_{\text{train}} \in [0, 2]$) and dissipative ($\delta_{\text{train}} > 2$) regimes form a robust "generalization plateau," maintaining high accuracy across the entire $\delta_{\text{test}}$ axis. Notably, the transition zone achieves the strongest overall generalization, effectively decoding inputs from all dynamical regimes.

This plateau is absent in recurrent controls (LSTMs, RNNs) and stateless MLPs (\textit{Last-T}), which show OOD failure (collapsed ridges) when tested away from their training diagonal. The \textit{Avg-Pool} MLP shows partial generalization but with significantly higher volatility (wider confidence bands in Figure~\ref{fig:1}a, see also Appendix Tables~\ref{tab:cross_enc_snn}--\ref{tab:cross_enc_rnn}). This comparison reveals a key distinction: \textit{Avg-Pool} achieves OOD robustness by collapsing input temporal variation before learning occurs, while SNNs induce invariance through the network learning process. The former provides a shortcut that discards information indiscriminately, which explains its higher performance volatility; the latter filters transient noise while retaining task-relevant features. Although LSTMs and RNNs preserve temporal autocorrelations, their poor performance suggests that generalization arises from alignment between signal dynamics and architecture, not temporal processing per se.

\textbf{Emergent Hierarchical Variance Reduction.}
\textit{How do SNNs translate these dynamical constraints into robust representations?} We hypothesize that dissipative constraints drive the network to suppress noise while amplifying stable features. To test this, we measured neural response stability via the coefficient of variation ($CV = \sigma/\mu$) of mean firing rates across test encodings. Figure~\ref{fig:1}f reveals a mechanistic insight: response variability (CV) strongly anti-correlates with OOD accuracy across the dynamical spectrum ($r=-0.96$; see Appendix Table~\ref{tab:cv_correlation_overall}). 

Furthermore, this stabilization emerges hierarchically across layers. As shown in Figure~\ref{fig:1}g, SNNs exhibit monotonic reduction with depth (L1 $\to$ L3) with the deepest layer showing the starkest contrast: when the training regime is under expansive dynamics, Layer 3 exhibits high fluctuating variance ($0.470 \pm 0.288$, see Appendix Table~\ref{tab:layer_cv_stats}), suggesting that the network memorizes transient trajectories of input signals. In contrast, dissipative constraints drive representation toward a tightly converged state ($0.307 \pm 0.051$), indicating invariant feature formation (Appendix Table~\ref{tab:layer_cv_stats}). 
This phenomenon is highly similar to the processing hierarchy observed in both standard deep networks~\cite{yosinski2014transferable} and the biological vision system (primate ventral stream), where raw, stimulus-specific volatility (akin to V1 responses) is progressively condensed into abstract representations (as in the inferotemporal cortex)~\cite{dicarlo2012does, akella2025deciphering}. However, previous works primarily attribute such invariance to prior spatial architecture designs (e.g., pooling or strided convolutions~\cite{yamins2014performance}); our results demonstrate that hierarchical robustness can emerge from intrinsic temporal dissipation alone.

\subsection{Experiment 2: Emergence of Structured Features}
\label{sec:structure}

The preceding findings naturally raise our next question: \textit{Does variance reduction spontaneously induce organized feature representations?} Classical theories posit sparse coding as an explicit optimization objective for cortical receptive fields~\cite{olshausen1996emergence, barlow1961possible}. We hypothesize instead that dissipative dynamics act as a fundamental constraint that naturally induces the invariant, structured features. In other words, representational structure is not the result of a designed objective but an emergent property of phase space compression.

To test this, we trained an SNN autoencoder to reconstruct static image patches from temporally encoded inputs. The architecture comprises a LIF bottleneck layer between a linear encoder ($W_{\text{enc}}$) and decoder ($W_{\text{dec}}$), where the decoder reconstructs from the temporal sum of spikes: $\mathbf{x}_{\text{recon}} = W_{\text{dec}} \sum_{t=1}^{T} S(t)$. Following established schemes~\cite{auge2021survey, guo2021neural}, we compared eight temporal encoding strategies on CIFAR-10 patches~\cite{krizhevsky2009learning}: \textit{Baseline} (static), \textit{Random} (temporal jitter), \textit{Linear} (rate coding), \textit{Poisson} (stochastic spiking), and four dynamical regimes (\textit{Expansive, Critical, Transition, Dissipative}). 

The network minimized a composite loss balancing \textit{reconstruction fidelity} and \textit{code sparsity} (see Appendix~\ref{app:exp2} for details). To evaluate learned representations, we measured the standard 
deviation of bottleneck receptive fields ($\sigma_{\text{RF}}$), which distinguishes structured filters from low-variance noise. Robustness was validated through: (1) varying the sparsity weight $\lambda$ (specifically including unconstrained $\lambda=0$); (2) scaling experiments to high-resolution STL-10 images~\cite{coates2011analysis}; and (3) testing dynamical universality across alternative systems (Lorenz~\cite{lorenz2017deterministic} and Thomas attractors~\cite{thomas1999deterministic}). Detailed experiment protocols are provided in Appendix~\ref{app:exp2}.

\begin{figure}[ht!]
    \centering
    \includegraphics[width=0.9\textwidth]{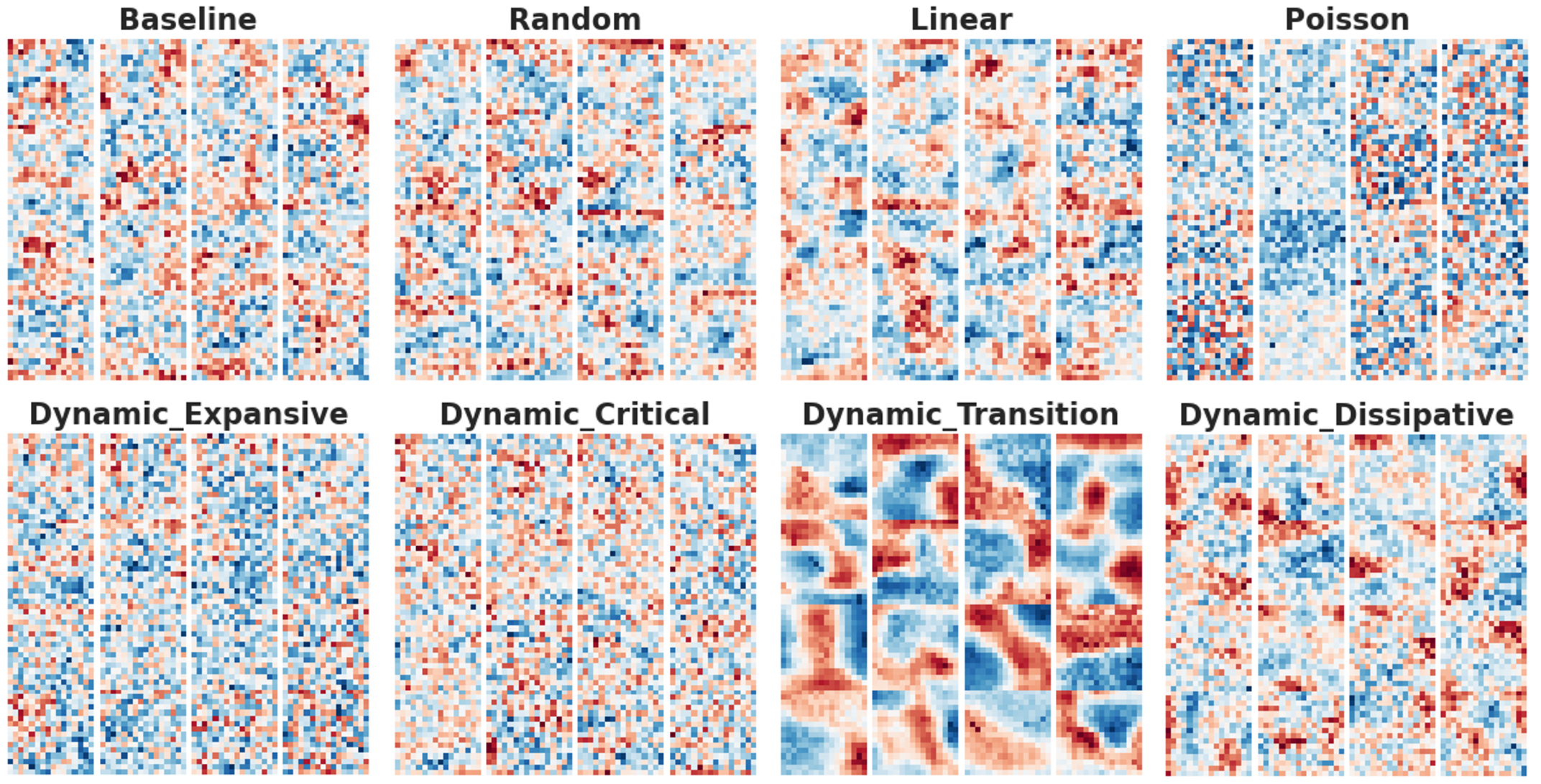}
    \caption{\textbf{Spontaneous emergence of structured receptive fields.} Visualization of learned features (receptive fields) at the network bottleneck. Red and blue pixels represent positive (excitatory) and negative (inhibitory) weights, respectively. Note that clear spatial antagonism emerges uniquely under \textit{Transition dynamics}, whereas other regimes yield unstructured noise.}
    \label{fig:2}
\end{figure}

\textbf{Structure Emergence under Dynamical Constraints}
Figure~\ref{fig:2} provides a qualitative visualization of feature emergence. \textit{Transition regime} ($\delta\approx2.0$) spontaneously produces receptive fields with clear spatial organization: localized, oriented filters resembling the biological orientation maps. In contrast, \textit{baseline}, \textit{random}, and \textit{expansive} conditions yield homogeneous, unstructured weight distributions resembling salt-and-pepper noise. Table~\ref{tab:3} quantifies these distinct organizations across three metrics:
\begin{itemize}
    \item \textit{Expansive dynamics ($\delta=-1.5$)}: Achieves excellent reconstruction ($0.0013$) but catastrophic sparsity ($1.72$), which means that dynamics amplify temporal variations, enabling memorization via high-rate firing without learning compressive features.
    \item \textit{Poisson encoding}: Demonstrates the opposite extreme with near-perfect sparsity ($10^{-5}$) but minimal structure ($0.049$). This indicates that sparse firing alone is insufficient; temporal stability is required to induce structural differentiation.
    \item \textit{Transition dynamics ($\delta\approx2.0$)}: Achieves optimal balance. Contractive phase space simultaneously reduces activation volume (sparsity $0.0009$), suppresses noise to enable clean filters, and amplifies stable features (structure $0.206$, $\mathbf{4.9\times}$ higher than alternatives).
\end{itemize}

\begin{table}[ht!]
\centering
\caption{Quantifying receptive field organization at the network bottleneck.}
\label{tab:3}
\begin{tabular}{lcccc}
\toprule
\textbf{Encoder} & \textbf{Dynamics ($\delta$)} & \textbf{Structure (RF Std)} & \textbf{Reconstruction Loss} & \textbf{Sparsity Loss} \\
\midrule
\textbf{Dynamic Transition} & \textbf{2.0} & \textbf{0.2058} & \textbf{0.003722} & \textbf{0.000868} \\
Dynamic Dissipative & 10.0 & 0.0502 & 0.003858 & 0.000804 \\
Linear & - & 0.0541 & 0.003506 & 0.002270 \\
Poisson & - & 0.0492 & 0.003908 & \textbf{0.000008} \\
Dynamic Expansive & -1.5 & 0.0435 & \textbf{0.001276} & 1.716117 \\
Baseline & - & 0.0417 & 0.003534 & 0.002529 \\
Dynamic Critical & 0.0 & 0.0367 & 0.003872 & 0.001046 \\
Random & - & 0.0341 & 0.003656 & 0.001896 \\
\bottomrule
\end{tabular}
\end{table}

Further analysis confirms that even without sparsity penalty ($\lambda=0$, Table~\ref{tab:ablation_lambda_all},  Appendix~\ref{app:exp2}), the \textit{Transition regime} still spontaneously induces structured representations ($\sigma_{\text{RF}} \approx 0.177$), whereas others collapse into noise. This structural emergence is robust: replication with Lorenz and Thomas attractors yields similar structured receptive fields exclusively within their dissipative regimes (Fig.~\ref{fig:universality_ablation}, Appendix~\ref{app:landscape}), confirming that this emergence is not specific to our oscillator system; a high-resolution parameter sweep reveals that this structural emergence persists as a stable ridge across the dynamical landscape (Fig.~\ref{fig:scan_grid}, Appendix~\ref{app:exp2}). 

\textbf{Mechanism}
\textit{What specific physical property drives this structural emergence?} Spectral analysis reveals that the transition regime ($\delta \approx2$) occupies a unique computational niche defined by a "High-Entropy, Low-Frequency" signature (Fig.~\ref{fig:spectral_mechanism}, Appendix~\ref{app:spectral}). This signature is scale-invariant across timescales ($T_{\max}$) and sampling densities ($N$), ensuring that the network receives information-rich structure within a slowly-varying envelope. By restricting input trajectories to a low spectral centroid, transition dynamics function as an intrinsic spectral shaper that aligns input complexity with the spectral bias of neural networks~\cite{rahaman2019spectral}, which also explains the asymmetric generalization observed in Experiment 1. These spectral properties confirm that feature organization arises solely from the dissipative dynamics itself, acting as a principle that induces structure more naturally than hand-crafted, explicit regularization.

Functionally, the observed structure (Figure~\ref{fig:2}) mirrors the spatial antagonism found in biological vision~\cite{olshausen1996emergence, ferster2000neural}. By suppressing high-frequency noise while preserving low-frequency structure, transition dynamics amplify spatiotemporal correlations. Such selective amplification biases the optimization process toward grouping co-varying features, yielding outcomes that are much like Hebbian association~\cite{hebb2005organization, barlow1961possible}, and it is driven by input temporal statistics rather than explicit plasticity rules. Conversely, high-frequency chaos in the expansive dynamics decorrelates inputs and prevents consolidation over time, while static baselines lack the necessary temporal continuity for structural binding. This suggests that dissipative constraints serve as a computational role beyond energy efficiency: by naturally filtering temporal correlations, they provide a mechanistic basis for structured connectivity~\cite{buzsaki2006rhythms}. More broadly, it provides a mechanistic account for biological sparse coding: under dissipative metabolic constraints, structured representations emerge not as a pre-programmed objective, but as a canonical solution to learning under temporal constraints~\cite{attwell2001energy}.

Beyond biological plausibility, this interpretation connects to recent work on diffusion dynamics~\cite{bonnaire2025diffusion}, which identifies a critical gap between memorization and generative abstraction. Our transition regime positions learning on the generative side of this divide: expansive dynamics with high-frequency chaos allow the network to memorize transient noise for rapid error reduction (the "memorization phase"), whereas the transition regime imposes a temporal information bottleneck, which makes the system discard transient variations and rely on the underlying generative invariants (Table~\ref{table:d1_oscillator_dynamics_transposed} and Fig.~\ref{fig:spectral_mechanism} in Appendix). This effectively implements Slow Feature Analysis~\cite{wiskott2002slow} through dynamics alone, reframing the "performance valley"~\cite {chen2025dynamical} not as a failure, but as the necessary cost of abstraction.

\subsection{Experiment 3: Zero-Shot Transfer to Unseen Physical Regimes}
\label{sec:robustness}

The preceding experiments established that dissipative constraints embedded in input encoding induce generalization at \textit{representational} (Exp 1) and \textit{structural} (Exp 2) levels. This naturally motivates our final test: if low-level invariance and structural emergence can be induced by temporal constraints, does this translate into high-level \textit{behavioral robustness} in sequential decision-making tasks? To address this, we adopted a zero-shot generalization paradigm in reinforcement learning (RL) tasks~\cite{packer2018assessing, kirk2023survey}: agents trained on a standard ``Easy'' environment were evaluated on progressively difficult physical regimes without retraining. Unlike approaches relying on extrinsic domain randomization~\cite{tobin2017domain, peng2018sim}, we posit that optimal dynamical constraints confer intrinsic robustness regardless of implementation level. We therefore designed a two-pathway validation:

\begin{itemize}
    \item \textbf{Encoding-Level.} To maintain consistency with our previous findings, we first applied the same encoding-level constraints to the \texttt{CartPole} task~\cite{towers2024gymnasium} within a zero-shot paradigm. Dynamical-encoded inputs were processed by SNNs or MLPs to directly isolate the role of temporal integration.
    
    \item \textbf{Architecture-Level.} While dynamical encoding probes how constraints influence learning, it relies on synthetically expanding input vectors into trajectories. However, RL environments provide inherently sequential signals. This enables a direct test of whether the constraint principle generalizes beyond artificial injection. We therefore varied the SNN's membrane leak parameter $\beta$, which governs intrinsic temporal integration. This parallels our encoding-level constraints: just as $\delta$ controls phase space contraction of input trajectories, $\beta$ controls information dissipation within the network's hidden states. We conducted systematic $\beta$-sweeps across \texttt{CartPole} (4D state, REINFORCE) and \texttt{LunarLander}~\cite{towers2024gymnasium} (8D state, PPO~\cite{schulman2017proximal}), comparing Leaky SNNs against MLP and LSTM baselines on raw state histories without external encoding (see Appendix~\ref{app:exp3_arch} for details).
\end{itemize}

Both validation approaches reveal consistent patterns (Tables~\ref{tab:rl_results}--\ref{tab:arch_comparison}): At the encoding level, \textit{SNN-Transition} ($\delta = 2.0$) achieved the lowest generalization gap (38.6 under fixed-budget; 42.4 with sufficient-training), significantly outperforming Expansive (87.6), Dissipative (53.1), and MLP Baseline (112.9). At the architecture level, intermediate dissipation ($\beta = 0.5$) consistently achieved optimal transfer: 90\% improvement over MLP on \texttt{CartPole}, and 46--50\% improvement over MLP and LSTM on \texttt{LunarLander}. Notably, SNNs exhibit non-monotonic performance across the $\beta$ spectrum: over-constrained regimes ($\beta \leq 0.3$) showed training instability, while unconstrained SNNs ($\beta = 1.0$) degraded generalization on complex tasks despite successful training. This cross-implementation convergence confirms that the optimal regime reflects a characteristic timescale balancing signal preservation against noise suppression, rather than implementation details.

\begin{table}[ht!]
\centering
\caption{Encoding-level generalization comparison (\texttt{CartPole}). Dynamical encoding transforms states into temporal trajectories with varying constraint strength ($\delta$). \textit{SNN-Transition} ($\delta=2.0$) achieves the lowest gap. Bold: best generalization (lowest gap).}
\label{tab:rl_results}
\resizebox{0.8\textwidth}{!}{
\begin{tabular}{llcccc}
\toprule
\textbf{Experiment} & \textbf{Agent Group} & \makecell{\textbf{Easy} \\ (Mean $\pm$ Std)} & \makecell{\textbf{V. Hard} \\ (Mean $\pm$ Std)} & \textbf{Avg. Gap} & \makecell{\textbf{Convergence} \\ (Median Eps.)} \\
\midrule
\multirow{7}{*}{\makecell[l]{Fixed Budget \\ (2000 Eps.)}} 
 & \textit{MLP Baseline} & $189.8 \pm 24.8$ & $37.3 \pm 39.0$ & 120.6 & --- \\
 & MLP Exp. ($\delta=-1.5$) & $179.5 \pm 31.8$ & $62.4 \pm 46.1$ & 89.3 & --- \\
 & MLP Tran. ($\delta=2.0$) & $192.8 \pm 11.4$ & $74.5 \pm 41.1$ & 79.8 & --- \\
 & MLP Diss. ($\delta=10.0$) & $178.6 \pm 44.0$ & $78.0 \pm 35.9$ & 56.5 & --- \\
 & SNN Exp. ($\delta=-1.5$) & $183.9 \pm 14.5$ & $67.0 \pm 34.5$ & 86.8 & --- \\
 & \textbf{SNN Tran.} ($\delta=2.0$) & $169.4 \pm 32.1$ & $\mathbf{106.1 \pm 29.6}$ & $\mathbf{38.6}$ & --- \\
 & SNN Diss. ($\delta=10.0$) & $165.0 \pm 29.5$ & $87.1 \pm 47.3$ & 56.1 & --- \\
\midrule
\multirow{7}{*}{\makecell[l]{Sufficient \\ Training}} 
 & \textit{MLP Baseline} & $199.2 \pm 1.5$ & $34.6 \pm 23.3$ & 112.9 & 510 \\
 & MLP Exp. ($\delta=-1.5$) & $198.7 \pm 1.9$ & $67.0 \pm 52.2$ & 105.4 & 970 \\
 & MLP Tran. ($\delta=2.0$) & $198.0 \pm 2.2$ & $80.6 \pm 52.6$ & 72.2 & 1360 \\
 & MLP Diss. ($\delta=10.0$) & $198.9 \pm 1.1$ & $67.2 \pm 41.3$ & 81.9 & 840 \\
 & SNN Exp. ($\delta=-1.5$) & $198.3 \pm 1.4$ & $70.3 \pm 48.3$ & 87.6 & 2870 \\
 & \textbf{SNN Tran.} ($\delta=2.0$) & $194.7 \pm 6.4$ & $\mathbf{126.3 \pm 34.2}$ & $\mathbf{42.4}$ & 2060 \\
 & SNN Diss. ($\delta=10.0$) & $193.6 \pm 8.2$ & $107.6 \pm 32.8$ & 53.1 & 2270 \\
\bottomrule
\end{tabular}%
}
\end{table}

\begin{table}[ht!]
\centering
\caption{Architecture-level generalization comparison. Varying membrane leak $\beta$ reveals a non-monotonic optimum. Note that over-constrained regimes ($\beta \le 0.3$) fail to converge reliably. Bold: best generalization (lowest gap).}
\label{tab:arch_comparison}
\resizebox{0.8\textwidth}{!}{
\begin{tabular}{llcccc}
\toprule
\textbf{Experiment} & \textbf{Agent Group} & \makecell{\textbf{Easy} \\ (Mean $\pm$ Std)} & \makecell{\textbf{V. Hard} \\ (Mean $\pm$ Std)} & \textbf{Avg. Gap} & \makecell{\textbf{Convergence} \\ (Median Eps.)} \\
\midrule
\multirow{8}{*}{CartPole} 
 & \textit{MLP (baseline)} & $199.0 \pm 2.0$ & $60.3 \pm 50.4$ & 138.7 & 100 \\
 & \textit{LSTM (baseline)} & $169.0 \pm 60.5$ & $127.0 \pm 65.2$ & 42.0 & 50 \\
 & Leaky SNN ($\beta=0.1$) & $8.4 \pm 0.1^\dagger$ & $15.4 \pm 0.3$ & --- & 1050 \\
 & Leaky SNN ($\beta=0.3$) & $10.9 \pm 4.9^\dagger$ & $20.0 \pm 8.9$ & --- & 1050 \\
 & \textbf{Leaky SNN ($\beta=0.5$)} & $195.5 \pm 2.7$ & $182.1 \pm 3.7$ & $\mathbf{13.4}$ & 2500 \\
 & Leaky SNN ($\beta=0.7$) & $198.2 \pm 1.5$ & $179.0 \pm 10.7$ & 19.2 & 1150 \\
 & Leaky SNN ($\beta=0.9$) & $196.9 \pm 2.0$ & $176.7 \pm 8.5$ & 20.2 & 950 \\
 & Leaky SNN ($\beta=1.0$) & $199.9 \pm 0.1$ & $186.4 \pm 6.4$ & 13.5 & 850 \\
\midrule
\multirow{8}{*}{LunarLander} 
 & \textit{MLP (baseline)} & $238.9 \pm 11.2$ & $58.6 \pm 40.5$ & 180.3 & 400 \\
 & \textit{LSTM (baseline)} & $220.0 \pm 14.7$ & $26.9 \pm 27.8$ & 193.1 & 200 \\
 & Leaky SNN ($\beta=0.1$) & $-79.3 \pm 289.3^\dagger$ & $-209.8 \pm 199.9$ & --- & 3600 \\
 & Leaky SNN ($\beta=0.3$) & $153.6 \pm 123.8^*$ & $28.5 \pm 24.2$ & 125.1 & 3600 \\
 & \textbf{Leaky SNN ($\beta=0.5$)} & $217.9 \pm 13.9$ & $121.4 \pm 18.0$ & $\mathbf{96.5}$ & 2400 \\
 & Leaky SNN ($\beta=0.7$) & $234.8 \pm 13.4$ & $122.4 \pm 19.8$ & 112.4 & 2000 \\
 & Leaky SNN ($\beta=0.9$) & $224.8 \pm 20.6$ & $96.4 \pm 38.9$ & 128.4 & 2000 \\
 & Leaky SNN ($\beta=1.0$) & $229.1 \pm 10.7$ & $125.0 \pm 21.9$ & 104.1 & 1600 \\
\bottomrule
\multicolumn{6}{l}{\footnotesize $^\dagger$Training failed to converge. $^*$High variance, partial convergence only.}
\end{tabular}%
}
\end{table}

Furthermore, the non-monotonic performance profile observed in Table~\ref{tab:arch_comparison} helps disentangle a critical confound: does generalization originate from the temporal constraint itself, or merely from the ``memory'' capacity inherent in SNN membrane dynamics~\cite{eshraghian2023training}? If memory alone drove generalization, performance should increase monotonically with information retention (high $\beta$). Instead, the observed degradation (or failure) at $\beta \approx 1.0$ indicates that intermediate dissipation implements a necessary temporal information bottleneck, filtering transient noise while preserving invariant structure. This interpretation is further reinforced by the evidence from additional validation with recurrent variants (\textit{RLeaky SNNs}, see Appendix~\ref{app:rleaky}): despite possessing additional memory pathways, these networks did not exhibit improved generalization. On the contrary, they displayed a significantly narrower stability window, failing to converge in unconstrained regimes ($\beta \approx 1.0$) where feedforward Leaky SNNs still functioned. This fragility suggests that excess memory without sufficient constraint destabilizes learning, revealing an embodied duality of the dynamical constraint, as it simultaneously governs both information retention (memory) and phase space contraction (forgetting).

In conclusion, these multi-level validations provide converging evidence for our central thesis: structured dissipation functions not as a limitation but as a temporal inductive bias, compelling networks to extract robust, invariant representations. The consistency of these results across our dual-pathway evaluation confirms that the generalization benefit is an intrinsic property of the constraint principle itself rather than the specificities of implementation.


\section{Discussion}

\textbf{Related Work}
Recent literature increasingly frames physical constraints not as computational limitations, but as essential inductive biases for intelligence. These constraints can be implemented at different system levels: through architectural design~\cite{hernandez2022thermodynamics}, learnable internal dynamics~\cite{zhang2024intrinsic}, or temporal hierarchies~\cite{cheng2020finite, moro2024role}. Our work advances this perspective by identifying temporal dissipation as a unique class of constraint, effective whether imposed via external input dynamics or internal state specification.

As demonstrated in Experiment 2, dissipative dynamics achieve spectral compression equivalent to Slow Feature Analysis~\cite{wiskott2002slow}, but as a physical consequence of phase space contraction rather than an explicit optimization objective. Critically, unlike spatial regularizers where dimensions are interchangeable, temporal constraints exploit an intrinsic asymmetry: the arrow of time transforms irreversible dissipation into a computational resource for extracting invariants. This positions temporal dissipation as a regularization mechanism parallel to spatial constraints: both impose structure that promotes generalization, but through distinct inductive biases. Our results indicate that temporal constraints alone can induce invariant representations required for out-of-distribution generalization~\cite{deng2022strong, goyal2022inductive}.

\textbf{From Generalization to Specialization}
Our experiments reveal another phenomenon: the transition regime exhibits slower convergence compared to expansive or strongly dissipative dynamics, yet achieves superior generalization. To rigorously ground this observation, we formalize this trade-off phenomenon through a complementary PAC-Bayes analysis~\cite{mcallester1998some} (Appendix~\ref{app:pacbayes}): by bounding the generalization error as a function of the posterior's alignment with the prior, this analysis reveals that the transition regime achieves the optimal balance between empirical fit (training error) and structural stability (Kullback–Leibler divergence) while exhibiting the most stable gradient norms (Appendix Table~\ref{tab:pacbayes_metrics}--\ref{tab:gradient_stats}). This provides a principled explanation for the observed cost of abstraction, i.e., the increased training effort: as evidenced by the delayed convergence in our behavioral tasks (Exp 3, Appendix Figure~\ref{fig:training_curves}), extracting robust invariants is inherently more computationally demanding than the rapid memorization of transient patterns. In other words, the very regime that appears suboptimal from an efficiency-centric view is, in fact, optimal for learning stable, transferable solutions.

This insight not only aligns with the "Edge of Chaos" theory~\cite{langton1990computation, bertschinger2004edge} but also explains its computational capacity by explicitly identifying state temporal dissipative constraints as an active regularization mechanism. Taken together, we envision a different landscape: rather than seeking a static optimum for all computation, the neural computation landscape should be viewed as a continuous dynamical spectrum, where the transition regime acts as an adaptability reservoir, maintaining the scale-invariant complexity required to extract robust invariants. Adjacent regimes serve distinct downstream objectives: shifting towards expansive dynamics ($\Sigma\lambda_i > 0$) maximizes discriminability for inference, or collapsing towards dissipative dynamics ($\Sigma\lambda_i \ll 0$) ensures stability and energy minimization. Therefore, this suggests that neural systems are better formalized as adaptive systems rather than static entities. The ``optimal'' configuration is not a fixed point but a regime to be traversed as task demands change, and neuronal properties become mutable states that are actively evolve to adapt to external environmental impacts and internal budgets. 

For machine learning, this necessitates a paradigm shift from architectural search to system dynamical regime navigation: robust generalization emerges from systems that modulate their temporal dynamics in response to task demands between retaining specific data fidelity and extracting generalizable understanding. The transition regime plays the role of a generalization optimum at a certain stage, which emerges from aligning the integration timescale to task-relevant invariants. This perspective offers a potential explanation for the apparent paradox that biological neural systems, despite vast memory capacity or energy efficiency, exhibit robust generalization: sparse coding and metabolic constraints may represent evolutionary solutions that naturally situate neural dynamics within the transition regime.

\textbf{Thermodynamic Perspective}
Why should dissipation, typically viewed as loss, confer 
computational principle? Our findings invite interpretation through non-equilibrium thermodynamics. From the perspective of the Second Law, an unconstrained expansive system naturally drifts toward maximal entropy, which is reflected in our experiments as isotropic noise and failure to generalize. The emergence of structured representations under dissipative constraints can be viewed as the formation of a \textit{dissipative structure}~\cite{prigogine1978time}: by continuously contracting phase space, the network actively resists entropic drift to maintain low-entropy invariant representations. 

This perspective provides an explanation for why temporal irreversibility is computationally vital. Unlike reversible Hamiltonian dynamics where information is conserved, intelligence systems require the arrow of time: it is the irreversible loss of transient information (forgetting) that drives the consolidation of stable structure (memory). Dissipative constraints enable networks to compress temporal variability into stable, generalizable representations, much as living systems maintain order through continuous entropy export~\cite{schrodinger2025life}. From this view, the metabolic constraints observed in biological neural systems are not mere resource limitations, but the thermodynamic conditions under which robust generalization becomes possible.

\textbf{Limitations and Future Directions}
While this work demonstrates the efficacy of dynamical constraints, our implementation primarily employs a fixed oscillator system as a controlled probe for mechanistic discovery. This represents only a static instantiation; performance reported herein likely represents a lower bound on the paradigm's potential. Similarly, although our architecture-level experiments ($\beta$-sweeps) demonstrate that the constraint principle generalizes beyond dynamical encoding, both implementations employ only fixed parameters for the network. Meanwhile, biological systems exhibit heterogeneous time constants across neural populations; exploring learned or adaptive $\beta$ distributions represents a natural extension. Importantly, the transition regime is defined not by a specific parameter values but a spectral signature: high entropy (preserved information complexity) combined with low dominant frequency (alignment with neural networks' spectral bias). Our scale-space analysis (Fig.~\ref{fig:spectral_mechanism}B) confirms this signature is scale-invariant, suggesting an intrinsic property of the dynamical regime rather than a tuning artifact. Nevertheless, verifying this regime-generality across alternative dynamical systems remains an important direction for future work.

\section{Conclusion}

This work identifies dissipative temporal dynamics as a distinct class of inductive bias, effective whether imposed through external or internal states. The critical transition regime does not merely filter noise; it reshapes the learning problem itself, compelling networks to extract invariants that static or unconstrained systems cannot access. These findings challenge a prevailing assumption in machine learning that removing constraints unlocks capability. Our results suggest the opposite relationship: the capacity for generalization may not only exist in architectural complexity or parameter scale, but in the dynamical regime through which information flows. This reframes the design objective from expanding model capacity to structuring the temporal dynamics through which abstraction learning occurs or where transient variation naturally separates from persistent structure. From this perspective, the metabolic pressures that confine biological neural dynamics to dissipative regimes may thus reflect not evolutionary compromise, but a principled solution to the generalization problem.

\section*{Acknowledgments}
This work was supported by the Georg Nemetschek Institute at the Technical University of Munich through the TUM GNI Postdoc Program. 
The source code and experimental data for this study are publicly available on GitHub at \url{https://github.com/chenxiachan/Constraint-breeds-generalization}.

\printbibliography[title={References}]
\end{refsection} 

\appendix
\section{Appendix}
\begin{refsection}
\addcontentsline{toc}{section}{Appendix}

\subsection*{Justification of Dynamical System \& Metric Selection}
\label{app:exp}
The central methodological challenge in studying temporal constraints as inductive biases is to isolate their effect from confounding factors. Classical chaotic systems couple their dynamical properties with specific geometric structures (e.g., the Lorenz butterfly, Rössler spiral, or Chua double-scroll), making it difficult to attribute computational effects to dynamics versus geometry.

We selected a modified duffing oscillator~\cite{kovacic2011duffing} for its \textit{minimal geometric complexity}. As a damped nonlinear oscillator, the Duffing system produces smooth, non-fractal phase trajectories. This geometric simplicity ensures that observed effects on generalization can be attributed to the global contraction rate ($\Sigma\lambda_i$) rather than attractor-specific topology. Concretely, this system offers three methodological advantages: (1) a single parameter $\delta$ permits continuous control of global phase space contraction, spanning from expansive ($\Sigma\lambda_i > 0$) to strongly dissipative ($\Sigma\lambda_i \ll 0$) regimes; (2) this parametric control decouples contraction rate from topological structure; and (3) its well-characterized bifurcation behavior~\cite{kovacic2011duffing} enables principled interpretation across the dynamical spectrum. Prior comparative work across classical chaotic attractors further supports this choice, having established that computational outcomes depend primarily on global Lyapunov structure rather than specific geometric manifolds~\cite{wolf1985determining, chen2025dynamical}. This makes the Duffing system an ideal standardized probe for our investigation. Each input feature $x_i$ initializes a three-dimensional dynamical system. The initial state $(x_0, y_0, z_0)$ is set as:
\begin{align}
    x_0 &= x_i \nonumber\\
    y_0 &= 0.2 \cdot x_i \\
    z_0 &= -x_i \nonumber
\end{align}
This linear mapping ensures that different feature values generate distinct trajectories, preserving discriminability while providing a consistent initialization scheme across the feature space.

The initial conditions evolve according to the coupled oscillator dynamics:
$$
\begin{aligned}
\dot{x} &= y \\
\dot{y} &= -\alpha x - \beta x^3 - \delta y + \gamma z \\
\dot{z} &= -\omega x - \delta z + \gamma xy
\end{aligned}
$$

where hyperparameters are fixed to $\alpha=2.0$, $\beta=0.1$, $\gamma=0.1$, $\omega=1.0$. The system evolves for a total time $T$ and is sampled at $N$ discrete timesteps, yielding a trajectory $\phi_\delta(x_i, t) \in \mathbb{R}^{N \times 3}$. For a $d$-dimensional input, the complete encoded representation is $\Phi_\delta(\mathbf{x}, t) \in \mathbb{R}^{d \times N \times 3}$. This parametrically controllable system allows for the continuous tuning of global phase space dynamics via a single parameter $\delta$, effectively decoupling the rate of phase space contraction from the topological structure of the attractor.

To characterize the physical behavior of this system, we systematically mapped its dynamical properties across the control spectrum. Table~\ref{table:d1_oscillator_dynamics_transposed} presents the relationship between $\delta$ and three key dynamical indicators: we employed the QR decomposition method~\cite{wolf1985determining} to calculate Lyapunov exponents. At each integration step, we computed the Jacobian matrix $J(\mathbf{x})$, updated an orthonormal basis via $Q_{\text{new}} = e^{J\Delta t}Q$, and performed periodic QR decomposition to accumulate eigenvalue contributions. The \textit{Maximum Lyapunov Exponent} ($\lambda_{\max}$) and \textit{Lyapunov Sum} ($\Sigma\lambda_i$) were obtained by normalizing accumulated values over total integration time. To characterize encoded trajectory quality, we computed \textit{Active Information Storage (AIS)}~\cite{lizier2012local}, which measures the mutual information between current and preceding states: $\text{AIS} = I(X_t; X_{t-1})$. Probability densities were estimated using equal-width histograms with Laplace smoothing, and the final metric was averaged across the three spatial dimensions. As evidenced by the data, the parameter $\delta$ provides monotonic control over the phase space contraction rate ($\Sigma\lambda_i$), spanning from the expansive regime ($\Sigma\lambda_i > 0$) to the strongly dissipative regime ($\Sigma\lambda_i \ll 0$).

\begin{table}[h!]
\centering
\caption{Dynamical property reference of the mixed oscillator system\cite{chen2025dynamical}. The parameter $\delta$ regulates global phase space dynamics. AIS (Active Information Storage) values in \textit{italics} highlight the information bottleneck in the transition regime ($\delta \approx 2.0$).}
\label{table:d1_oscillator_dynamics_transposed}
\resizebox{\textwidth}{!}{%
\begin{tabular}{@{}l|ccccccccccccccccc@{}}
\toprule
\textbf{$\delta$} & \textbf{-1.5} & \textbf{-1.0} & \textbf{-0.6} & \textbf{-0.3} & \textbf{-0.15} & \textbf{0} & \textbf{0.15} & \textbf{0.3} & \textbf{0.6} & \textbf{1.0} & \textbf{1.5} & \textbf{2.0} & \textbf{2.5} & \textbf{4.0} & \textbf{5.0} & \textbf{7.0} & \textbf{10.0} \\ \midrule
$\lambda_{\max}$ & 1.17 & 0.79 & 0.44 & 0.24 & 0.16 & 0.08 & 0 & -0.07 & -0.22 & -0.41 & -0.70 & -0.88 & -1.15 & -0.55 & -0.42 & -0.29 & -0.20 \\
$\Sigma\lambda_i$ & 3 & 2 & 1.2 & 0.6 & 0.3 & 0 & -0.3 & -0.6 & -1.2 & -2 & -3 & -4 & -5 & -8 & -10 & -14 & -20 \\
AIS & 2.99 & 2.95 & 3.05 & 2.80 & 3.02 & 2.97 & 2.99 & 2.99 & 2.66 & 2.15 & \textit{1.62} & \textit{1.06} & \textit{1.09} & 2.07 & 2.44 & 2.84 & 3.08 \\ \bottomrule
\end{tabular}
}
\end{table}

These empirical metrics justify our definition of the \textit{Transition} regime ($\delta \approx 2.0$) as a unique computational state. While the Lyapunov sum decreases monotonically with $\delta$, the AIS exhibits a distinct non-monotonic behavior, reaching a global minimum in the region $\delta \in [1.5, 2.5]$. Specifically, at $\delta=2.0$, the system enters a state characterized simultaneously by weak dissipation ($\Sigma\lambda_i = -4$), indicating phase space contraction, and a critical information bottleneck. In this state, the AIS drops to its lowest level ($\approx 1.06$), which is significantly lower than both the expansive ($\approx 2.99$) and strongly dissipative ($\approx 3.08$) extremes. This sharp dip in information storage capacity corresponds to a state of minimal short-term memory retention. This characteristic is consistent with a filtering mechanism that suppresses transient noise, corroborating the emergence of robust, invariant features described in the main text.

For our experiments, we span two complementary configurations to demonstrate the robustness of the principle:
\begin{itemize}
    \item \textit{High-Resolution Mapping (Exp 1):} To rigorously map the generalization landscape, we employed a dense sampling regime ($T=4.0, N=30$). This setting minimizes numerical drift to ensure precise topological mapping.
    
    \item \textit{Efficient Application (Exp 2 \& 3):} For downstream tasks (Receptive Field Learning and RL), we adopted a computationally efficient sparse regime ($T=8.0, N=5$). This configuration accommodates the stricter computational budget inherent in these iterative training loops while retaining sufficient dynamical complexity.
\end{itemize}

As evidenced by our scale-space analysis (Fig.~\ref{fig:spectral_mechanism}B), the mechanism is regime-dependent rather than parameter-specific; the critical ``low-frequency, high-entropy'' signature of the transition regime persists across varying observation scales ($N$) and evolution times ($T_{max}$), the mechanism governed by $\delta$ remains invariant: tuning $\delta$ continuously traverses the spectrum from expansive ($\delta < 0$) to dissipative ($\delta \gg 0$) dynamics, with the transition regime ($\delta \approx 2.0$) consistently emerging as the critical state for generalization.

\subsection{Experiment 1: Cross-Encoding Generalization Details}
\label{app:exp1}
This appendix provides detailed experimental setups and supplementary analyses for the representational level experiments described in Section~\ref{sec:generalization}. 

\textbf{Dataset and Setup.} We use the \texttt{sklearn.datasets.load\_digits} dataset (64 features, 10 classes)~\cite{optical_recognition_of_handwritten_digits_80,scikit-learn}, split into 70\% training, 15\% validation, and 15\% testing. Input features $\mathbf{x}$ are first normalized using StandardScaler. Networks are trained on data encoded with one $\delta_{\text{train}}$ value, then tested on 12 different encodings: $\delta_{\text{test}} \in \{-1.5, -1.0, -0.3, 0.0, 0.3, 1.0, 1.5, 2.0, 2.5, 5.0, 7.0, 10.0\}$. The detailed quantitative generalization matrices (Mean $\pm$ Std \%) for all architectures are provided in Tables~\ref{tab:cross_enc_snn} - \ref{tab:cross_enc_rnn}.

\textbf{Training.} LIF neurons use learnable thresholds and membrane decay $\beta=0.95$. Training uses surrogate gradients (\texttt{surrogate.fast\_sigmoid} with \texttt{slope=25})~\cite{eshraghian2023training} with the Adam optimizer (\texttt{lr=1e-4})~\cite{kingma2014adam} for 200 epochs, with early stopping (\texttt{patience=10}) based on validation accuracy.

\begin{table*}[htp!]
\centering
\caption{Cross-encoding generalization (mean $\pm$ std \%) for the SNN model. Rows indicate $\delta_{\text{train}}$, columns indicate $\delta_{\text{test}}$. Diagonal (in-distribution) results are bolded.}
\label{tab:cross_enc_snn}
\resizebox{\textwidth}{!}{%
\begin{tabular}{@{}l|cccccccccccc@{}}
\toprule
\textbf{Train $\delta$} & \multicolumn{12}{c}{\textbf{Test $\delta$}} \\
& \turnbox{0}{\textbf{-1.5}} & \turnbox{0}{\textbf{-1.0}} & \turnbox{0}{\textbf{-0.3}} & \turnbox{0}{\textbf{0.0}} & \turnbox{0}{\textbf{0.3}} & \turnbox{0}{\textbf{1.0}} & \turnbox{0}{\textbf{1.5}} & \turnbox{0}{\textbf{2.0}} & \turnbox{0}{\textbf{2.5}} & \turnbox{0}{\textbf{5.0}} & \turnbox{0}{\textbf{7.0}} & \turnbox{0}{\textbf{10.0}} \\
\midrule
\textbf{-1.5} & \textbf{92.9 $\pm$ 1.9} & 92.0 $\pm$ 2.0 & 56.1 $\pm$ 10.7 & 38.7 $\pm$ 11.6 & 29.0 $\pm$ 9.6 & 21.2 $\pm$ 6.9 & 18.6 $\pm$ 6.7 & 17.9 $\pm$ 6.5 & 18.0 $\pm$ 6.6 & 17.6 $\pm$ 7.2 & 17.9 $\pm$ 7.2 & 17.6 $\pm$ 6.6 \\
\textbf{-1.0} & 90.9 $\pm$ 5.9 & \textbf{91.7 $\pm$ 6.3} & 71.4 $\pm$ 5.8 & 51.0 $\pm$ 7.9 & 36.1 $\pm$ 8.4 & 21.1 $\pm$ 6.8 & 17.3 $\pm$ 6.3 & 16.1 $\pm$ 6.0 & 15.1 $\pm$ 5.6 & 14.4 $\pm$ 5.0 & 14.1 $\pm$ 5.0 & 14.1 $\pm$ 4.5 \\
\textbf{-0.3} & 85.8 $\pm$ 2.8 & 90.3 $\pm$ 2.2 & \textbf{95.2 $\pm$ 1.0} & 94.5 $\pm$ 1.0 & 92.4 $\pm$ 1.9 & 87.8 $\pm$ 4.1 & 85.3 $\pm$ 5.6 & 84.4 $\pm$ 6.4 & 82.3 $\pm$ 6.7 & 79.3 $\pm$ 7.1 & 77.9 $\pm$ 8.6 & 76.8 $\pm$ 8.2 \\
\textbf{0.0} & 81.4 $\pm$ 6.1 & 86.6 $\pm$ 3.6 & 94.3 $\pm$ 1.8 & \textbf{94.6 $\pm$ 1.8} & 94.9 $\pm$ 1.5 & 92.8 $\pm$ 1.9 & 92.3 $\pm$ 2.3 & 91.2 $\pm$ 2.6 & 90.9 $\pm$ 2.2 & 89.2 $\pm$ 2.6 & 89.0 $\pm$ 2.6 & 88.5 $\pm$ 2.3 \\
\textbf{0.3} & 81.0 $\pm$ 2.6 & 86.2 $\pm$ 1.8 & 93.5 $\pm$ 1.6 & 94.8 $\pm$ 1.7 & \textbf{94.7 $\pm$ 1.8} & 94.2 $\pm$ 1.7 & 93.4 $\pm$ 2.3 & 93.5 $\pm$ 2.8 & 93.1 $\pm$ 2.6 & 92.2 $\pm$ 2.1 & 91.9 $\pm$ 2.5 & 90.9 $\pm$ 2.4 \\
\textbf{1.0} & 79.6 $\pm$ 2.8 & 84.7 $\pm$ 2.4 & 92.6 $\pm$ 1.8 & 94.4 $\pm$ 1.5 & 94.3 $\pm$ 1.7 & \textbf{94.5 $\pm$ 1.9} & 94.2 $\pm$ 1.8 & 93.9 $\pm$ 2.2 & 93.6 $\pm$ 2.2 & 93.3 $\pm$ 2.3 & 92.9 $\pm$ 2.8 & 92.7 $\pm$ 2.4 \\
\textbf{1.5} & 72.5 $\pm$ 4.7 & 78.6 $\pm$ 4.4 & 90.3 $\pm$ 1.9 & 92.9 $\pm$ 1.7 & 94.4 $\pm$ 1.8 & 94.8 $\pm$ 1.7 & \textbf{94.6 $\pm$ 2.0} & 94.5 $\pm$ 2.1 & 94.4 $\pm$ 1.8 & 93.5 $\pm$ 2.1 & 93.2 $\pm$ 2.0 & 93.0 $\pm$ 2.1 \\
\textbf{2.0} & 70.9 $\pm$ 8.1 & 76.9 $\pm$ 8.5 & 90.3 $\pm$ 2.6 & 92.8 $\pm$ 1.7 & 93.8 $\pm$ 1.9 & 94.3 $\pm$ 1.9 & 94.3 $\pm$ 1.8 & \textbf{94.2 $\pm$ 2.0} & 93.9 $\pm$ 2.4 & 93.6 $\pm$ 2.3 & 93.4 $\pm$ 2.0 & 93.3 $\pm$ 2.0 \\
\textbf{2.5} & 68.5 $\pm$ 6.9 & 74.4 $\pm$ 6.1 & 90.0 $\pm$ 2.2 & 93.2 $\pm$ 1.8 & 93.7 $\pm$ 1.4 & 94.4 $\pm$ 1.4 & 94.8 $\pm$ 1.2 & 94.6 $\pm$ 1.1 & \textbf{94.8 $\pm$ 1.1} & 94.6 $\pm$ 1.4 & 94.4 $\pm$ 1.5 & 93.9 $\pm$ 1.4 \\
\textbf{5.0} & 65.1 $\pm$ 6.9 & 71.1 $\pm$ 5.8 & 86.3 $\pm$ 3.2 & 89.9 $\pm$ 2.3 & 91.5 $\pm$ 2.3 & 93.4 $\pm$ 1.7 & 93.8 $\pm$ 1.6 & 94.3 $\pm$ 1.8 & 94.1 $\pm$ 1.8 & \textbf{94.7 $\pm$ 1.3} & 94.7 $\pm$ 1.8 & 94.2 $\pm$ 1.7 \\
\textbf{7.0} & 66.9 $\pm$ 8.9 & 71.7 $\pm$ 7.6 & 86.0 $\pm$ 4.8 & 89.9 $\pm$ 2.9 & 91.5 $\pm$ 2.1 & 93.3 $\pm$ 1.4 & 93.6 $\pm$ 1.2 & 93.7 $\pm$ 1.9 & 94.2 $\pm$ 1.6 & 94.5 $\pm$ 1.6 & \textbf{94.5 $\pm$ 1.4} & 94.6 $\pm$ 1.5 \\
\textbf{10.0} & 69.1 $\pm$ 6.3 & 74.2 $\pm$ 5.5 & 85.7 $\pm$ 3.0 & 89.6 $\pm$ 2.4 & 91.2 $\pm$ 2.1 & 92.5 $\pm$ 2.3 & 92.9 $\pm$ 2.4 & 93.5 $\pm$ 2.2 & 93.5 $\pm$ 1.9 & 94.1 $\pm$ 1.7 & 94.5 $\pm$ 1.9 & \textbf{94.6 $\pm$ 2.1} \\
\bottomrule
\end{tabular}%
}
\end{table*}

\begin{table*}[htp!]
\centering
\caption{Cross-encoding generalization (mean $\pm$ std \%) for the MLP (Avg-Pool) model. Rows indicate $\delta_{\text{train}}$, columns indicate $\delta_{\text{test}}$. Diagonal (in-distribution) results are bolded.}
\label{tab:cross_enc_mlp_avg}
\resizebox{\textwidth}{!}{%
\begin{tabular}{@{}l|cccccccccccc@{}}
\toprule
\textbf{Train $\delta$} & \multicolumn{12}{c}{\textbf{Test $\delta$}} \\
& \turnbox{0}{\textbf{-1.5}} & \turnbox{0}{\textbf{-1.0}} & \turnbox{0}{\textbf{-0.3}} & \turnbox{0}{\textbf{0.0}} & \turnbox{0}{\textbf{0.3}} & \turnbox{0}{\textbf{1.0}} & \turnbox{0}{\textbf{1.5}} & \turnbox{0}{\textbf{2.0}} & \turnbox{0}{\textbf{2.5}} & \turnbox{0}{\textbf{5.0}} & \turnbox{0}{\textbf{7.0}} & \turnbox{0}{\textbf{10.0}} \\
\midrule
\textbf{-1.5} & \textbf{94.9 $\pm$ 1.3} & 94.8 $\pm$ 1.4 & 67.1 $\pm$ 9.4 & 39.2 $\pm$ 13.4 & 23.7 $\pm$ 9.7 & 15.8 $\pm$ 5.1 & 12.0 $\pm$ 3.8 & 10.4 $\pm$ 3.0 & 8.9 $\pm$ 2.9 & 6.3 $\pm$ 4.0 & 5.6 $\pm$ 4.1 & 4.9 $\pm$ 4.1 \\
\textbf{-1.0} & 95.0 $\pm$ 1.8 & \textbf{95.0 $\pm$ 1.7} & 93.2 $\pm$ 1.6 & 87.9 $\pm$ 2.5 & 69.3 $\pm$ 6.9 & 16.5 $\pm$ 4.0 & 5.3 $\pm$ 2.0 & 2.0 $\pm$ 1.2 & 1.4 $\pm$ 1.2 & 0.6 $\pm$ 0.7 & 0.6 $\pm$ 0.8 & 0.5 $\pm$ 0.6 \\
\textbf{-0.3} & 93.8 $\pm$ 1.9 & 94.1 $\pm$ 1.9 & \textbf{94.5 $\pm$ 1.9} & 93.9 $\pm$ 2.0 & 92.7 $\pm$ 2.2 & 19.7 $\pm$ 9.0 & 2.5 $\pm$ 3.2 & 0.9 $\pm$ 1.5 & 0.5 $\pm$ 0.9 & 0.2 $\pm$ 0.3 & 0.2 $\pm$ 0.3 & 0.2 $\pm$ 0.3 \\
\textbf{0.0} & 90.6 $\pm$ 8.0 & 90.7 $\pm$ 7.9 & 90.8 $\pm$ 8.5 & \textbf{91.1 $\pm$ 8.2} & 91.1 $\pm$ 7.6 & 90.6 $\pm$ 7.1 & 88.3 $\pm$ 6.5 & 83.8 $\pm$ 5.9 & 64.1 $\pm$ 7.5 & 0.5 $\pm$ 0.9 & 0.2 $\pm$ 0.2 & 0.1 $\pm$ 0.2 \\
\textbf{0.3} & 91.9 $\pm$ 2.5 & 92.1 $\pm$ 2.6 & 92.5 $\pm$ 3.0 & 92.8 $\pm$ 2.9 & \textbf{92.7 $\pm$ 2.7} & 92.0 $\pm$ 2.3 & 91.2 $\pm$ 1.9 & 88.6 $\pm$ 1.9 & 74.7 $\pm$ 10.6 & 0.4 $\pm$ 0.6 & 0.2 $\pm$ 0.4 & 0.1 $\pm$ 0.3 \\
\textbf{1.0} & 90.7 $\pm$ 3.3 & 90.8 $\pm$ 3.3 & 91.0 $\pm$ 3.3 & 91.3 $\pm$ 3.4 & 91.5 $\pm$ 3.6 & \textbf{91.8 $\pm$ 3.3} & 91.7 $\pm$ 3.4 & 91.7 $\pm$ 3.4 & 91.6 $\pm$ 3.5 & 91.7 $\pm$ 3.6 & 91.7 $\pm$ 3.5 & 91.7 $\pm$ 3.5 \\
\textbf{1.5} & 91.0 $\pm$ 2.4 & 91.0 $\pm$ 2.5 & 91.3 $\pm$ 2.6 & 92.1 $\pm$ 2.6 & 92.1 $\pm$ 2.9 & 92.1 $\pm$ 2.8 & \textbf{92.0 $\pm$ 2.9} & 91.9 $\pm$ 2.9 & 91.9 $\pm$ 2.9 & 91.8 $\pm$ 3.0 & 91.8 $\pm$ 2.9 & 91.9 $\pm$ 2.9 \\
\textbf{2.0} & 86.5 $\pm$ 14.5 & 86.6 $\pm$ 14.8 & 86.9 $\pm$ 14.1 & 87.0 $\pm$ 13.9 & 87.2 $\pm$ 14.0 & 87.6 $\pm$ 13.8 & 87.7 $\pm$ 13.9 & \textbf{87.6 $\pm$ 14.0} & 87.6 $\pm$ 14.0 & 87.6 $\pm$ 14.2 & 87.4 $\pm$ 14.3 & 87.3 $\pm$ 14.2 \\
\textbf{2.5} & 88.5 $\pm$ 9.9 & 88.6 $\pm$ 9.7 & 88.9 $\pm$ 9.6 & 89.2 $\pm$ 9.4 & 89.5 $\pm$ 9.7 & 89.4 $\pm$ 9.5 & 89.3 $\pm$ 9.4 & 89.4 $\pm$ 9.2 & \textbf{89.2 $\pm$ 9.4} & 89.2 $\pm$ 9.5 & 89.1 $\pm$ 9.3 & 89.0 $\pm$ 9.3 \\
\textbf{5.0} & 91.3 $\pm$ 2.7 & 91.3 $\pm$ 2.6 & 92.0 $\pm$ 2.2 & 92.0 $\pm$ 1.9 & 92.0 $\pm$ 1.8 & 92.0 $\pm$ 1.9 & 92.1 $\pm$ 2.0 & 92.1 $\pm$ 2.1 & 92.1 $\pm$ 2.1 & \textbf{92.1 $\pm$ 2.2} & 92.1 $\pm$ 2.1 & 92.1 $\pm$ 2.1 \\
\textbf{7.0} & 87.2 $\pm$ 11.8 & 87.4 $\pm$ 11.5 & 87.4 $\pm$ 10.8 & 87.7 $\pm$ 10.5 & 87.7 $\pm$ 10.3 & 88.1 $\pm$ 10.3 & 88.2 $\pm$ 10.3 & 88.3 $\pm$ 10.2 & 88.3 $\pm$ 10.0 & 88.4 $\pm$ 10.3 & \textbf{88.5 $\pm$ 10.3} & 88.5 $\pm$ 10.3 \\
\textbf{10.0} & 92.3 $\pm$ 2.2 & 92.3 $\pm$ 2.3 & 92.4 $\pm$ 2.1 & 92.4 $\pm$ 1.9 & 92.5 $\pm$ 1.9 & 92.9 $\pm$ 1.7 & 92.9 $\pm$ 1.8 & 92.9 $\pm$ 1.8 & 92.9 $\pm$ 1.7 & 93.2 $\pm$ 1.7 & 93.3 $\pm$ 1.7 & \textbf{93.3 $\pm$ 1.7} \\
\bottomrule
\end{tabular}%
}
\end{table*}

\begin{table*}[htp!]
\centering
\caption{Cross-encoding generalization (mean $\pm$ std \%) for the MLP (T-last) model. Rows indicate $\delta_{\text{train}}$, columns indicate $\delta_{\text{test}}$. Diagonal (in-distribution) results are bolded.}
\label{tab:cross_enc_mlp_tlast}
\resizebox{\textwidth}{!}{%
\begin{tabular}{@{}l|cccccccccccc@{}}
\toprule
\textbf{Train $\delta$} & \multicolumn{12}{c}{\textbf{Test $\delta$}} \\
& \turnbox{0}{\textbf{-1.5}} & \turnbox{0}{\textbf{-1.0}} & \turnbox{0}{\textbf{-0.3}} & \turnbox{0}{\textbf{0.0}} & \turnbox{0}{\textbf{0.3}} & \turnbox{0}{\textbf{1.0}} & \turnbox{0}{\textbf{1.5}} & \turnbox{0}{\textbf{2.0}} & \turnbox{0}{\textbf{2.5}} & \turnbox{0}{\textbf{5.0}} & \turnbox{0}{\textbf{7.0}} & \turnbox{0}{\textbf{10.0}} \\
\midrule
\textbf{-1.5} & \textbf{92.3 $\pm$ 2.3} & 90.2 $\pm$ 2.6 & 21.8 $\pm$ 2.3 & 11.0 $\pm$ 1.5 & 9.4 $\pm$ 2.2 & 10.0 $\pm$ 2.7 & 10.2 $\pm$ 2.2 & 10.5 $\pm$ 1.6 & 10.5 $\pm$ 1.6 & 11.3 $\pm$ 1.7 & 11.4 $\pm$ 1.9 & 11.4 $\pm$ 1.9 \\
\textbf{-1.0} & 94.4 $\pm$ 1.9 & \textbf{94.9 $\pm$ 1.8} & 89.1 $\pm$ 3.3 & 68.4 $\pm$ 9.9 & 34.1 $\pm$ 12.7 & 13.1 $\pm$ 5.7 & 11.1 $\pm$ 3.2 & 9.4 $\pm$ 1.7 & 9.3 $\pm$ 1.7 & 13.3 $\pm$ 5.7 & 15.3 $\pm$ 7.1 & 18.8 $\pm$ 8.3 \\
\textbf{-0.3} & 94.1 $\pm$ 2.2 & 94.8 $\pm$ 1.9 & \textbf{95.0 $\pm$ 2.3} & 94.4 $\pm$ 2.3 & 93.2 $\pm$ 2.2 & 45.8 $\pm$ 15.1 & 13.1 $\pm$ 4.1 & 6.9 $\pm$ 2.8 & 7.0 $\pm$ 3.2 & 44.1 $\pm$ 14.3 & 72.5 $\pm$ 10.2 & 85.2 $\pm$ 5.5 \\
\textbf{0.0} & 93.4 $\pm$ 1.6 & 94.0 $\pm$ 1.5 & 94.1 $\pm$ 1.8 & \textbf{94.2 $\pm$ 1.8} & 93.9 $\pm$ 2.0 & 62.3 $\pm$ 12.7 & 14.2 $\pm$ 6.8 & 7.4 $\pm$ 4.2 & 8.2 $\pm$ 3.8 & 54.4 $\pm$ 12.8 & 82.9 $\pm$ 4.8 & 89.8 $\pm$ 2.1 \\
\textbf{0.3} & 0.1 $\pm$ 0.2 & 0.3 $\pm$ 0.4 & 86.3 $\pm$ 11.4 & 88.7 $\pm$ 10.9 & \textbf{89.3 $\pm$ 10.3} & 86.7 $\pm$ 9.9 & 50.0 $\pm$ 4.7 & 12.9 $\pm$ 4.1 & 7.1 $\pm$ 2.7 & 31.6 $\pm$ 5.9 & 75.7 $\pm$ 8.7 & 86.0 $\pm$ 9.8 \\
\textbf{1.0} & 0.0 $\pm$ 0.0 & 0.0 $\pm$ 0.1 & 80.1 $\pm$ 16.0 & 81.0 $\pm$ 14.8 & 81.3 $\pm$ 14.3 & \textbf{81.5 $\pm$ 14.8} & 70.6 $\pm$ 16.2 & 15.6 $\pm$ 6.7 & 6.3 $\pm$ 3.8 & 36.3 $\pm$ 13.4 & 77.4 $\pm$ 17.1 & 80.6 $\pm$ 16.8 \\
\textbf{1.5} & 0.1 $\pm$ 0.4 & 0.1 $\pm$ 0.4 & 0.1 $\pm$ 0.4 & 0.9 $\pm$ 0.7 & 71.9 $\pm$ 18.9 & 72.7 $\pm$ 18.9 & \textbf{73.0 $\pm$ 19.5} & 67.6 $\pm$ 17.7 & 6.8 $\pm$ 3.9 & 0.2 $\pm$ 0.3 & 0.2 $\pm$ 0.4 & 0.2 $\pm$ 0.4 \\
\textbf{2.0} & 1.3 $\pm$ 1.7 & 1.1 $\pm$ 1.4 & 1.3 $\pm$ 2.6 & 3.7 $\pm$ 3.4 & 24.7 $\pm$ 12.6 & 26.0 $\pm$ 11.7 & 24.8 $\pm$ 12.0 & \textbf{22.4 $\pm$ 12.5} & 7.8 $\pm$ 3.5 & 2.1 $\pm$ 3.9 & 2.0 $\pm$ 3.9 & 2.0 $\pm$ 3.9 \\
\textbf{2.5} & 25.4 $\pm$ 4.9 & 23.2 $\pm$ 4.9 & 6.3 $\pm$ 3.7 & 4.7 $\pm$ 4.0 & 4.8 $\pm$ 4.2 & 6.3 $\pm$ 4.7 & 7.5 $\pm$ 4.6 & 10.4 $\pm$ 1.4 & \textbf{10.6 $\pm$ 1.0} & 10.8 $\pm$ 1.6 & 8.5 $\pm$ 4.1 & 7.9 $\pm$ 4.7 \\
\textbf{5.0} & 88.1 $\pm$ 3.3 & 87.7 $\pm$ 3.2 & 87.9 $\pm$ 3.5 & 74.4 $\pm$ 5.0 & 0.2 $\pm$ 0.2 & 0.1 $\pm$ 0.2 & 0.2 $\pm$ 0.3 & 0.8 $\pm$ 1.0 & 12.1 $\pm$ 3.9 & \textbf{89.6 $\pm$ 2.6} & 89.4 $\pm$ 2.8 & 89.2 $\pm$ 2.9 \\
\textbf{7.0} & 87.9 $\pm$ 6.4 & 87.4 $\pm$ 6.3 & 87.5 $\pm$ 6.5 & 70.0 $\pm$ 10.3 & 0.3 $\pm$ 0.5 & 0.3 $\pm$ 0.5 & 0.4 $\pm$ 0.7 & 1.4 $\pm$ 2.3 & 10.2 $\pm$ 1.5 & 88.9 $\pm$ 6.0 & \textbf{89.1 $\pm$ 6.4} & 89.0 $\pm$ 6.3 \\
\textbf{10.0} & 91.0 $\pm$ 2.7 & 90.7 $\pm$ 2.6 & 91.1 $\pm$ 2.6 & 72.0 $\pm$ 5.0 & 0.4 $\pm$ 0.4 & 0.2 $\pm$ 0.2 & 0.4 $\pm$ 0.4 & 1.4 $\pm$ 0.9 & 10.8 $\pm$ 3.1 & 92.1 $\pm$ 2.7 & 92.9 $\pm$ 2.9 & \textbf{92.7 $\pm$ 2.8} \\
\bottomrule
\end{tabular}%
}
\end{table*}

\begin{table*}[]
\centering
\caption{Cross-encoding generalization (mean $\pm$ std \%) for the LSTM model. Rows indicate $\delta_{\text{train}}$, columns indicate $\delta_{\text{test}}$. Diagonal (in-distribution) results are bolded.}
\label{tab:cross_enc_lstm}
\resizebox{\textwidth}{!}{%
\begin{tabular}{@{}l|cccccccccccc@{}}
\toprule
\textbf{Train $\delta$} & \multicolumn{12}{c}{\textbf{Test $\delta$}} \\
& \turnbox{0}{\textbf{-1.5}} & \turnbox{0}{\textbf{-1.0}} & \turnbox{0}{\textbf{-0.3}} & \turnbox{0}{\textbf{0.0}} & \turnbox{0}{\textbf{0.3}} & \turnbox{0}{\textbf{1.0}} & \turnbox{0}{\textbf{1.5}} & \turnbox{0}{\textbf{2.0}} & \turnbox{0}{\textbf{2.5}} & \turnbox{0}{\textbf{5.0}} & \turnbox{0}{\textbf{7.0}} & \turnbox{0}{\textbf{10.0}} \\
\midrule
\textbf{-1.5} & \textbf{87.6 $\pm$ 1.6} & 87.6 $\pm$ 2.1 & 79.3 $\pm$ 3.9 & 72.2 $\pm$ 4.2 & 64.8 $\pm$ 4.9 & 51.3 $\pm$ 6.3 & 44.0 $\pm$ 7.6 & 37.3 $\pm$ 8.0 & 32.3 $\pm$ 7.3 & 20.4 $\pm$ 6.4 & 16.7 $\pm$ 5.7 & 14.1 $\pm$ 5.1 \\
\textbf{-1.0} & 87.4 $\pm$ 2.8 & \textbf{89.6 $\pm$ 2.4} & 83.0 $\pm$ 3.7 & 69.7 $\pm$ 7.0 & 54.7 $\pm$ 8.1 & 33.4 $\pm$ 7.1 & 25.8 $\pm$ 6.6 & 20.9 $\pm$ 5.8 & 18.0 $\pm$ 4.2 & 11.4 $\pm$ 3.0 & 9.2 $\pm$ 2.9 & 8.1 $\pm$ 2.6 \\
\textbf{-0.3} & 71.4 $\pm$ 3.8 & 82.1 $\pm$ 3.2 & \textbf{94.0 $\pm$ 2.2} & 89.0 $\pm$ 1.9 & 71.1 $\pm$ 4.3 & 25.6 $\pm$ 6.3 & 16.1 $\pm$ 6.7 & 12.7 $\pm$ 5.3 & 10.7 $\pm$ 4.8 & 9.7 $\pm$ 3.7 & 10.0 $\pm$ 3.6 & 10.4 $\pm$ 2.8 \\
\textbf{0.0} & 60.2 $\pm$ 6.5 & 72.9 $\pm$ 5.4 & 92.4 $\pm$ 2.0 & \textbf{93.5 $\pm$ 1.5} & 85.8 $\pm$ 2.9 & 32.8 $\pm$ 9.0 & 16.4 $\pm$ 7.4 & 12.3 $\pm$ 7.2 & 10.7 $\pm$ 6.3 & 9.0 $\pm$ 5.3 & 9.0 $\pm$ 5.5 & 9.0 $\pm$ 5.5 \\
\textbf{0.3} & 7.5 $\pm$ 3.2 & 12.3 $\pm$ 4.9 & 86.2 $\pm$ 4.2 & 93.3 $\pm$ 1.5 & \textbf{93.4 $\pm$ 1.6} & 68.8 $\pm$ 6.1 & 35.7 $\pm$ 7.0 & 19.8 $\pm$ 5.1 & 12.5 $\pm$ 4.1 & 4.0 $\pm$ 1.5 & 2.7 $\pm$ 1.2 & 2.4 $\pm$ 1.1 \\
\textbf{1.0} & 5.1 $\pm$ 2.4 & 7.0 $\pm$ 3.4 & 60.6 $\pm$ 5.1 & 78.5 $\pm$ 4.3 & 86.5 $\pm$ 4.0 & \textbf{92.1 $\pm$ 1.6} & 80.1 $\pm$ 3.3 & 42.3 $\pm$ 11.0 & 24.2 $\pm$ 9.3 & 7.5 $\pm$ 4.7 & 5.3 $\pm$ 4.3 & 3.9 $\pm$ 3.6 \\
\textbf{1.5} & 3.0 $\pm$ 1.3 & 4.7 $\pm$ 2.5 & 46.5 $\pm$ 7.9 & 57.1 $\pm$ 8.5 & 64.5 $\pm$ 6.6 & 87.5 $\pm$ 3.4 & \textbf{91.9 $\pm$ 2.4} & 82.4 $\pm$ 7.1 & 57.0 $\pm$ 8.3 & 18.8 $\pm$ 6.0 & 13.3 $\pm$ 5.0 & 9.8 $\pm$ 4.7 \\
\textbf{2.0} & 8.3 $\pm$ 3.4 & 10.9 $\pm$ 3.9 & 36.2 $\pm$ 6.1 & 41.4 $\pm$ 9.4 & 49.1 $\pm$ 10.6 & 73.1 $\pm$ 3.7 & 86.0 $\pm$ 2.2 & \textbf{89.4 $\pm$ 2.4} & 87.4 $\pm$ 2.1 & 48.7 $\pm$ 7.9 & 34.4 $\pm$ 6.8 & 25.6 $\pm$ 6.8 \\
\textbf{2.5} & 50.3 $\pm$ 9.4 & 50.8 $\pm$ 9.4 & 41.4 $\pm$ 9.8 & 47.4 $\pm$ 11.7 & 62.4 $\pm$ 10.5 & 82.0 $\pm$ 5.8 & 85.6 $\pm$ 4.4 & 86.2 $\pm$ 5.0 & \textbf{86.4 $\pm$ 4.8} & 83.3 $\pm$ 4.5 & 81.0 $\pm$ 4.5 & 78.0 $\pm$ 4.7 \\
\textbf{5.0} & 46.9 $\pm$ 5.6 & 46.8 $\pm$ 6.5 & 32.7 $\pm$ 6.2 & 42.9 $\pm$ 7.1 & 60.5 $\pm$ 8.2 & 82.7 $\pm$ 3.5 & 87.1 $\pm$ 2.2 & 89.3 $\pm$ 2.3 & 90.2 $\pm$ 2.2 & \textbf{90.2 $\pm$ 2.5} & 89.8 $\pm$ 2.7 & 89.4 $\pm$ 2.8 \\
\textbf{7.0} & 47.3 $\pm$ 6.2 & 45.9 $\pm$ 6.9 & 30.6 $\pm$ 5.9 & 37.9 $\pm$ 7.5 & 55.3 $\pm$ 8.5 & 80.0 $\pm$ 3.2 & 85.0 $\pm$ 3.0 & 87.1 $\pm$ 3.4 & 88.2 $\pm$ 3.4 & 89.9 $\pm$ 3.1 & \textbf{89.9 $\pm$ 3.4} & 89.8 $\pm$ 3.7 \\
\textbf{10.0} & 50.8 $\pm$ 6.3 & 49.6 $\pm$ 5.7 & 33.9 $\pm$ 7.5 & 39.4 $\pm$ 8.5 & 53.9 $\pm$ 9.3 & 79.6 $\pm$ 4.6 & 85.0 $\pm$ 4.0 & 86.9 $\pm$ 3.9 & 88.1 $\pm$ 3.5 & 89.3 $\pm$ 3.2 & 89.2 $\pm$ 3.2 & \textbf{89.3 $\pm$ 3.3} \\
\bottomrule
\end{tabular}%
}
\end{table*}

\begin{table*}[]
\centering
\caption{Cross-encoding generalization (Mean $\pm$ Std \%) for the RNN model. Rows indicate $\delta_{\text{train}}$, columns indicate $\delta_{\text{test}}$. Diagonal (in-distribution) results are bolded.}
\label{tab:cross_enc_rnn}
\resizebox{\textwidth}{!}{%
\begin{tabular}{@{}l|cccccccccccc@{}}
\toprule
\textbf{Train $\delta$} & \multicolumn{12}{c}{\textbf{Test $\delta$}} \\
& \turnbox{0}{\textbf{-1.5}} & \turnbox{0}{\textbf{-1.0}} & \turnbox{0}{\textbf{-0.3}} & \turnbox{0}{\textbf{0.0}} & \turnbox{0}{\textbf{0.3}} & \turnbox{0}{\textbf{1.0}} & \turnbox{0}{\textbf{1.5}} & \turnbox{0}{\textbf{2.0}} & \turnbox{0}{\textbf{2.5}} & \turnbox{0}{\textbf{5.0}} & \turnbox{0}{\textbf{7.0}} & \turnbox{0}{\textbf{10.0}} \\
\midrule
\textbf{-1.5} & \textbf{83.5 $\pm$ 3.9} & 83.7 $\pm$ 3.3 & 52.0 $\pm$ 7.7 & 23.9 $\pm$ 6.0 & 15.5 $\pm$ 4.4 & 16.2 $\pm$ 6.1 & 17.2 $\pm$ 5.9 & 18.1 $\pm$ 5.1 & 18.0 $\pm$ 5.5 & 18.3 $\pm$ 5.9 & 17.9 $\pm$ 6.2 & 17.3 $\pm$ 6.3 \\
\textbf{-1.0} & 88.1 $\pm$ 2.4 & \textbf{90.6 $\pm$ 2.1} & 75.1 $\pm$ 3.2 & 45.5 $\pm$ 6.6 & 25.4 $\pm$ 3.7 & 15.0 $\pm$ 3.3 & 14.2 $\pm$ 4.6 & 15.2 $\pm$ 5.5 & 16.9 $\pm$ 7.0 & 20.0 $\pm$ 6.2 & 21.5 $\pm$ 6.3 & 22.2 $\pm$ 6.2 \\
\textbf{-0.3} & 68.7 $\pm$ 4.8 & 79.8 $\pm$ 4.2 & \textbf{94.1 $\pm$ 1.9} & 88.0 $\pm$ 2.8 & 65.4 $\pm$ 4.8 & 14.9 $\pm$ 5.2 & 7.2 $\pm$ 2.3 & 7.6 $\pm$ 3.0 & 9.0 $\pm$ 3.1 & 17.7 $\pm$ 5.7 & 23.3 $\pm$ 5.1 & 28.7 $\pm$ 6.0 \\
\textbf{0.0} & 52.3 $\pm$ 5.3 & 65.6 $\pm$ 6.8 & 89.7 $\pm$ 2.7 & \textbf{92.8 $\pm$ 2.8} & 86.4 $\pm$ 3.7 & 22.1 $\pm$ 6.8 & 8.3 $\pm$ 3.2 & 6.8 $\pm$ 2.7 & 7.4 $\pm$ 3.4 & 16.9 $\pm$ 5.0 & 23.2 $\pm$ 7.4 & 29.2 $\pm$ 8.2 \\
\textbf{0.3} & 1.9 $\pm$ 2.5 & 3.3 $\pm$ 2.8 & 72.1 $\pm$ 4.1 & 91.2 $\pm$ 2.5 & \textbf{93.0 $\pm$ 2.5} & 67.0 $\pm$ 8.4 & 25.0 $\pm$ 10.2 & 9.3 $\pm$ 6.1 & 5.1 $\pm$ 4.3 & 3.7 $\pm$ 3.8 & 6.9 $\pm$ 5.5 & 10.6 $\pm$ 6.6 \\
\textbf{1.0} & 1.2 $\pm$ 1.0 & 2.0 $\pm$ 1.5 & 44.6 $\pm$ 6.8 & 71.9 $\pm$ 4.9 & 79.7 $\pm$ 3.4 & \textbf{87.6 $\pm$ 3.3} & 63.2 $\pm$ 4.4 & 14.0 $\pm$ 5.5 & 5.5 $\pm$ 2.8 & 1.8 $\pm$ 1.4 & 1.2 $\pm$ 1.2 & 0.9 $\pm$ 0.8 \\
\textbf{1.5} & 1.0 $\pm$ 1.3 & 0.9 $\pm$ 0.9 & 19.5 $\pm$ 5.6 & 36.9 $\pm$ 12.3 & 42.4 $\pm$ 12.6 & 60.5 $\pm$ 11.1 & \textbf{78.2 $\pm$ 6.0} & 28.5 $\pm$ 8.6 & 11.5 $\pm$ 6.7 & 4.0 $\pm$ 3.8 & 2.8 $\pm$ 2.8 & 2.3 $\pm$ 2.7 \\
\textbf{2.0} & 2.4 $\pm$ 2.2 & 2.8 $\pm$ 2.2 & 7.0 $\pm$ 4.4 & 9.5 $\pm$ 5.3 & 10.4 $\pm$ 8.2 & 18.3 $\pm$ 7.6 & 33.7 $\pm$ 7.2 & \textbf{63.7 $\pm$ 14.1} & 25.1 $\pm$ 7.6 & 4.9 $\pm$ 1.8 & 3.8 $\pm$ 1.5 & 2.8 $\pm$ 1.3 \\
\textbf{2.5} & 24.7 $\pm$ 5.3 & 21.5 $\pm$ 5.2 & 3.8 $\pm$ 2.5 & 0.9 $\pm$ 1.3 & 1.8 $\pm$ 2.1 & 7.4 $\pm$ 5.8 & 17.2 $\pm$ 9.5 & 25.8 $\pm$ 13.2 & \textbf{26.7 $\pm$ 15.4} & 23.0 $\pm$ 7.8 & 21.7 $\pm$ 6.0 & 20.0 $\pm$ 4.5 \\
\textbf{5.0} & 36.9 $\pm$ 10.0 & 33.0 $\pm$ 8.8 & 5.1 $\pm$ 1.5 & 1.2 $\pm$ 0.8 & 1.3 $\pm$ 1.4 & 7.5 $\pm$ 3.5 & 24.3 $\pm$ 10.6 & 47.4 $\pm$ 13.7 & 67.9 $\pm$ 12.4 & \textbf{79.2 $\pm$ 7.3} & 78.8 $\pm$ 6.8 & 77.9 $\pm$ 6.7 \\
\textbf{7.0} & 46.6 $\pm$ 5.6 & 43.4 $\pm$ 5.6 & 6.9 $\pm$ 2.9 & 1.4 $\pm$ 1.0 & 1.1 $\pm$ 0.8 & 6.2 $\pm$ 2.9 & 25.9 $\pm$ 7.4 & 55.6 $\pm$ 8.7 & 75.5 $\pm$ 3.1 & 85.9 $\pm$ 2.5 & \textbf{86.1 $\pm$ 3.1} & 86.1 $\pm$ 3.1 \\
\textbf{10.0} & 42.7 $\pm$ 6.5 & 39.8 $\pm$ 6.1 & 6.6 $\pm$ 3.2 & 0.8 $\pm$ 0.6 & 0.7 $\pm$ 0.7 & 3.5 $\pm$ 2.8 & 14.4 $\pm$ 7.6 & 40.4 $\pm$ 10.8 & 62.1 $\pm$ 12.6 & 83.6 $\pm$ 4.4 & 85.8 $\pm$ 3.3 & \textbf{86.7 $\pm$ 3.1} \\
\bottomrule
\end{tabular}%
}
\end{table*}

\textbf{Overall Correlation Across All Train-Test Pairs.}
Table~\ref{tab:cv_correlation_overall} presents the Pearson correlation coefficients between layer-wise CV and OOD accuracy across all 1,440 samples (12 train encodings $\times$ 12 test encodings $\times$ 10 runs). All layers exhibit significant negative correlations, with correlation strength decreasing hierarchically from Layer 1 to Layer 3. This pattern suggests that early-layer stability is most predictive of generalization performance. Statistical significance was assessed using two-tailed tests with $\alpha = 0.05$.


\begin{table}[htp!]
\centering
\caption{Correlation between layer-wise CV and OOD accuracy across all train-test pairs (n=1440)}
\label{tab:cv_correlation_overall}
\begin{tabular}{lccc}
\toprule
\textbf{Layer} & \textbf{Pearson's r} & \textbf{p-value} & \textbf{Significance} \\
\midrule
Layer 1 & -0.783 & 4.96e-299 & *** \\
Layer 2 & -0.602 & 1.88e-142 & *** \\
Layer 3 & -0.215 & 1.66e-16 & *** \\
\bottomrule
\end{tabular}
\begin{tablenotes}
\small
\item Note: CV = Coefficient of Variation ($\sigma/\mu$). Negative correlations indicate that lower CV (more stable firing) is associated with higher OOD accuracy. Significance levels: *** p < 0.001, ** p < 0.01, * p < 0.05, n.s. = not significant.
\end{tablenotes}
\end{table}


\begin{table}[htp!]
\centering
\caption{Comparison of layer-wise CV across dynamical regimes. }
\label{tab:layer_cv_stats}
\begin{tabular}{llc}
\toprule
\textbf{Training Regime} & \textbf{Layer} & \textbf{Mean CV ($\mu \pm \sigma$)} \\
\midrule
Expansive ($\delta < 0$) & Layer 1 & 1.214 $\pm$ 0.062 \\
($n=240$) & Layer 2 & 0.716 $\pm$ 0.114 \\
 & Layer 3 & \textbf{0.470 $\pm$ 0.288} \\ 
\midrule
Critical / Transition ($\delta \in [0, 5]$) & Layer 1 & 0.723 $\pm$ 0.071 \\
($n=960$) & Layer 2 & 0.524 $\pm$ 0.055 \\
 & Layer 3 & 0.377 $\pm$ 0.084 \\
\midrule
Dissipative ($\delta > 5$) & Layer 1 & 0.673 $\pm$ 0.036 \\
($n=240$) & Layer 2 & 0.475 $\pm$ 0.026 \\
 & Layer 3 & \textbf{0.307 $\pm$ 0.051} \\ 
\bottomrule
\end{tabular}
\end{table}

\subsection{Experiment 2: Unsupervised Receptive Field Learning Details}
This appendix provides detailed experimental setups and supplementary analyses for the structural level experiments described in Section~\ref{sec:structure}.

\label{app:exp2}
\textbf{Dataset and Setup.} 
The model for this experiment is a spiking neural network (SNN) autoencoder. Its architecture consists of an encoder linear layer ($W_{\text{enc}} \in \mathbb{R}^{128 \times d}$), a recurrent Leaky Integrate-and-Fire (LIF) hidden layer with 128 learnable thresholds and membrane decay $\beta=0.95$ neurons~\cite{eshraghian2023training}, and a decoder linear layer ($W_{\text{dec}} \in \mathbb{R}^{d \times 128}$). The sparse autoencoder minimizes a composite loss balancing reconstruction fidelity and code sparsity, inspired by the sparse coding hypothesis of visual cortex~\cite{olshausen1996emergence, barlow1961possible}:
$$
\mathcal{L} = ||\mathbf{x}_{\text{target}} - \mathbf{x}_{\text{recon}}||_2^2 + \lambda ||\mathbf{z}_{\text{spikes}}||_1
$$
where $\mathbf{z}_{\text{spikes}} = \sum_{t=1}^{T} S(t)$ is the total spike count vector. We quantify receptive field (RF) organization and training dynamics via three metrics: \textit{Reconstruction Loss} ($||\mathbf{x}_{\text{target}} - \mathbf{x}_{\text{recon}}||_2^2$), \textit{Sparsity Loss} ($||\mathbf{z}||_1$), and our primary metric for structural organization, \textit{RF Standard Deviation} ($\sigma_{\text{RF}}$). 

\textit{Metric Definition:} $\sigma_{\text{RF}} = \sqrt{\frac{1}{N} \sum (D_{ij} - \bar{D})^2}$ quantifies the variance of learned weights, serving as a proxy for feature differentiation (structured filters have high variance, noise has low variance), where $D$ is the learned dictionary $W_{\text{enc}}$ and $\bar{D}$ is the global mean weight.

The learned receptive fields correspond to the rows of $W_{\text{enc}}$. The decoder reconstructs the static image from the temporal sum of spikes: $\mathbf{x}_{\text{recon}} = W_{\text{dec}} \sum_{t=1}^{T} S(t)$. For temporally encoded inputs, the network processes the signal sequentially. 

The dataset consists of 5,000 random $16 \times 16$ grayscale patches extracted from CIFAR-10 images (converted from $32 \times 32$ RGB via standard luminance formula)~\cite{krizhevsky2009learning}. Patches are normalized to the [0,1] range. We systematically compare eight input encoding strategies, which are grouped by their temporal characteristics~\cite{auge2021survey, guo2021neural}:
\begin{itemize}
    \item \textit{Baseline (static)}: $\mathbf{x}_{\text{in}} = \mathbf{x}_{\text{target}}$. The static patch is fed at $t=0$.
    \item \textit{Random (temporal jitter)}: Each pixel's value is perturbed by $\mathcal{N}(0, 0.1)$ noise across $T$ steps.
    \item \textit{Linear (rate coding)}: Pixel intensity $x_i$ linearly modulates firing rate, $r_i(t) = x_i \cdot t/T$.
    \item \textit{Poisson (stochastic spiking)}: Pixels generate Poisson spike trains with rate $\lambda_i = x_i \cdot f_{\text{max}}$.
    \item \textit{Dynamic Expansive ($\delta=-1.5$)}; \textit{Critical ($\delta=0.0$)}; \textit{Dissipative ($\delta=10.0$)}; \textit{Transition ($\delta=2.0$)}
\end{itemize}

\begin{figure}[ht]
  \begin{center}
    \centerline{\includegraphics[width=\columnwidth]{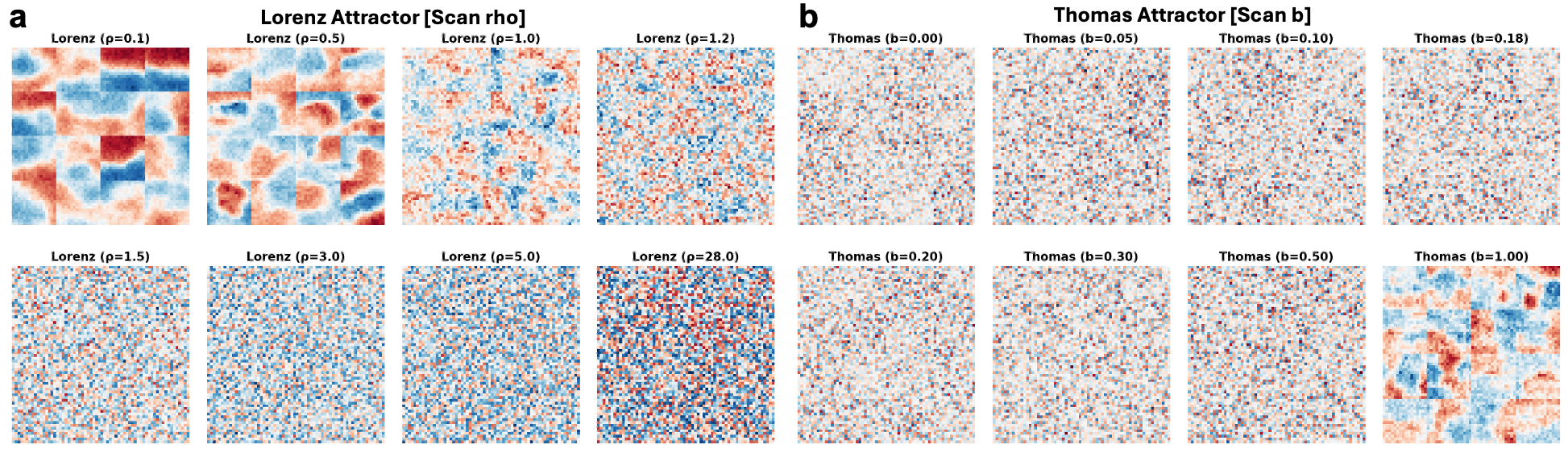}}
    \caption{\textbf{Dynamical Universality Validation.} We verify the constraint principle across two alternative dynamical systems to disentangle the effect of dissipation from specific equation geometries. 
    \textbf{(a)} Lorenz System: Structure emerges in the stable regime (low $\rho$) but collapses into noise as the system enters the classical chaotic regime ($\rho=28$). 
    \textbf{(b)} Thomas Attractor: At low damping ($b < 0.2$), hyper-chaotic dynamics prevent learning. As dissipative constraint increases ($b \to 1.0$), forcing the system into a transition regime, structured features spontaneously emerge. 
    Consistent results across polynomial (Lorenz) and sinusoidal (Thomas) non-linearities confirm that phase space contraction is the universal driver of structural generalization.}
    \label{fig:universality_ablation}
  \end{center}
\end{figure}

\begin{figure}[ht!]
    \centering
    \includegraphics[width=\textwidth]{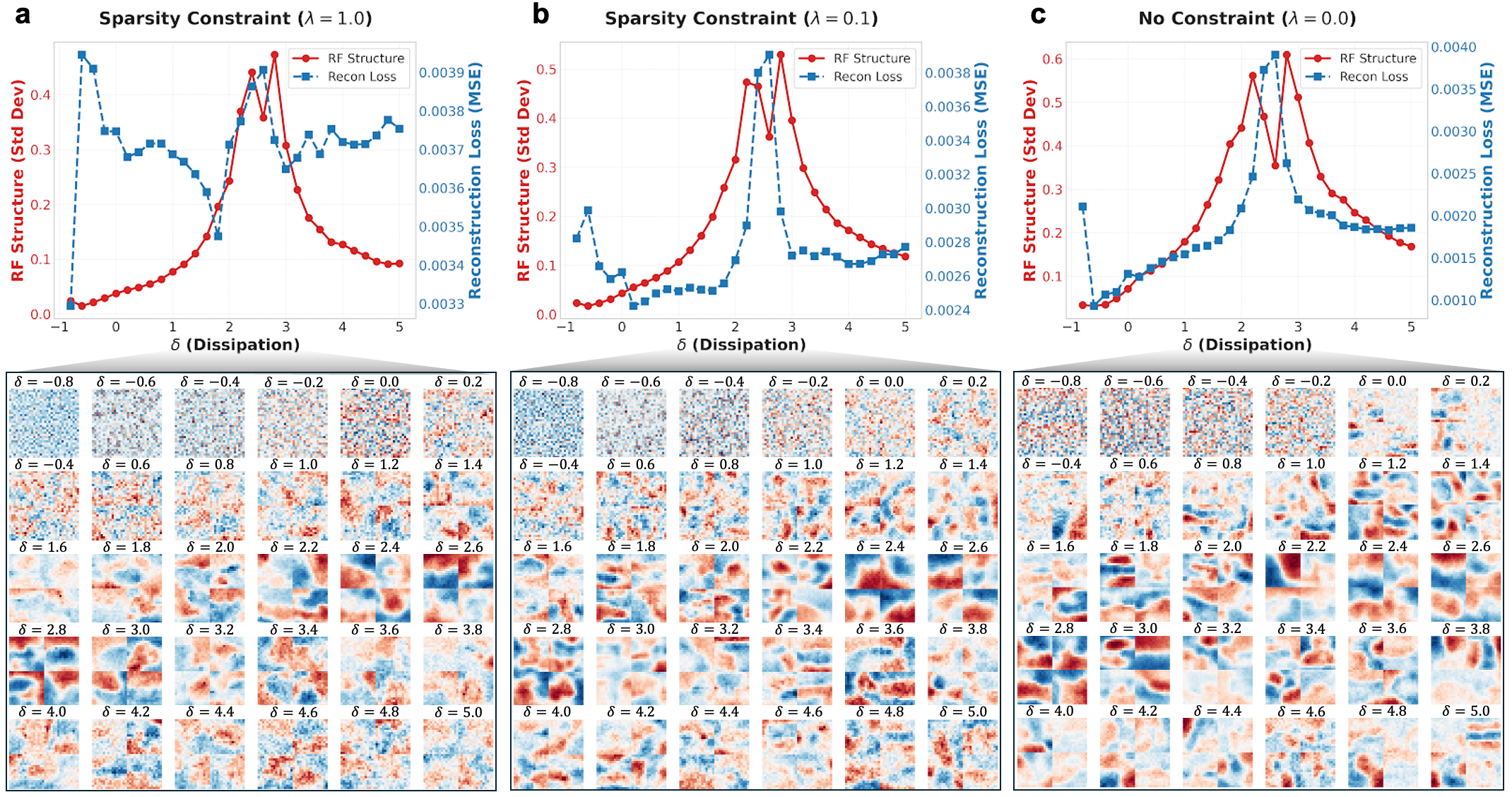} 
    \caption{\textbf{Intrinsic Dynamics and the Emergence of Structure.} 
    We analyze the impact of dynamical constraints ($\delta$) on feature organization across three sparsity regimes: \textbf{(a)} Standard constraint ($\lambda=1.0$), \textbf{(b)} Weak constraint ($\lambda=0.1$), and \textbf{(c)} No constraint ($\lambda=0.0$). 
    While a "structural ridge" ($\delta \in [1, 3]$) persists across all regimes, the peak structural organization \textit{increases} as the explicit sparsity constraint decreases (from $\sim 0.47$ at $\lambda=1.0$ to $\sim 0.61$ at $\lambda=0.0$). This trend is visually corroborated by the receptive field visualizations (bottom panels), where the diversity and clarity of structured filters visibly expand as the external constraint is relaxed. This demonstrates that the transition dynamics alone serve as a novel intrinsic inductive bias compared to external regularization.}
    \label{fig:scan_grid}
\end{figure}

Training uses the Adam optimizer~\cite{kingma2014adam} with a learning rate of $1 \times 10^{-3}$ using a batch size of 64. The encoder weight matrix $\mathbf{W}_{\text{enc}} \in \mathbb{R}^{128 \times 256}$ maps flattened image patches to the bottleneck layer. Each row of $\mathbf{W}_{\text{enc}}$ represents the synaptic weights connecting all 256 input pixels to a single bottleneck neuron. To visualize the learned receptive field of neuron $i$, we reshape its weight vector $\mathbf{w}_i \in \mathbb{R}^{256}$ back to the original spatial dimensions ($16 \times 16$). Figure~\ref{fig:2}, ~\ref{fig:universality_ablation}, and ~\ref{fig:scan_grid} display grids of such receptive fields, where red and blue pixels indicate positive (excitatory) and negative (inhibitory) connection weights, respectively. The standard deviation of these weights ($\sigma_{\text{RF}}$) quantifies structural differentiation: high variance indicates spatially organized filters, while low variance indicates homogeneous noise.

\textbf{Robustness Ablations.}
To validate the robustness of the findings in Table~\ref{tab:3} and ensure the structural emergence is not an artifact of specific hyperparameters, we design three follow-up validations on:
\begin{enumerate}
    \item \textit{Sparsity Weight ($\lambda$)}: We extended the evaluation to include $\lambda=0$ (no explicit sparsity constraint) alongside $\lambda \in \{0.1, 0.5, 1.0, 3.0\}$. As detailed in Table~\ref{tab:ablation_lambda_all}, the Transition regime ($\delta=2.0$) uniquely demonstrates spontaneous structural emergence even at $\lambda=0$ ($\sigma_{\text{RF}} \approx 0.177$), whereas other regimes collapse into unstructured noise ($\sigma_{\text{RF}} < 0.05$). Across all non-zero weights, the Transition regime consistently yields the highest Structural Score.
    \item \textit{Resolution}: We repeated experiments on STL-10 ($96 \times 96$ patches)~\cite{coates2011analysis}. Table~\ref{tab:ablation_stl10} confirms the findings hold for a different dataset and higher-resolution images.
    \item \textit{Dynamical Universality}: To confirm that the principle is not an artifact of the duffing oscillator, we replicated the receptive field learning experiment using two distinct systems: the Lorenz attractor (atmospheric convection)~\cite{lorenz2017deterministic} and the Thomas cyclically symmetric attractor (lattice dynamics)~\cite{thomas1999deterministic}.
\end{enumerate}

For the \textit{dynamical universality} validation, we selected systems that allow independent control of dissipation versus dynamical complexity:

\begin{itemize}
\item \textit{Lorenz System} (varying entropy at fixed dissipation): Defined by $\dot{x} = \sigma(y-x)$, $\dot{y} = x(\rho-z)-y$, $\dot{z} = xy-\beta z$, with standard parameters $\sigma=10, \beta=8/3$. We scanned the Rayleigh number $\rho \in [0.1, 28]$. In this setup, the phase space contraction rate remains constant ($\nabla \cdot F = -(\sigma+\beta+1)$) while $\rho$ increases the topological entropy.
\item \textit{Thomas Attractor} (varying dissipation at fixed topology): Defined by $\dot{x} = \sin(y) - bx$, $\dot{y} = \sin(z) - by$, $\dot{z} = \sin(x) - bz$. The parameter $b$ linearly controls the global energy dissipation ($\nabla \cdot F = -3b$).
\end{itemize}

As illustrated in Figure~\ref{fig:universality_ablation}, both systems exhibit a distinct phase transition dependent on their dynamical regimes: Lorenz system successfully learns structured representations in the stable regime ($\rho < 1.5$) but degrades into unstructured noise as the system enters the classical chaotic regime (e.g., $\rho=28$). This confirms that chaotic complexity alone without sufficient dissipative constraint is destructive. Similarly, the Thomas system demonstrates that at low damping ($b < 0.2$, hyper-chaotic), the network fails to learn structure, whereas strong dissipation ($b \to 1.0$) forces the system into a constrained regime where structured features spontaneously emerge.

\begin{table*}[]
\centering
\caption{Robustness of receptive field organization across varying sparsity weights ($\lambda$). 
\textit{Dynamic Transition} ($\delta=2.0$) regime consistently produces the highest structure (RF Std) across all weights, demonstrating that the emergence of structure is robust to changes in the sparsity objective. Bold values indicate the optimal result for each metric within each $\lambda$ group.}
\label{tab:ablation_lambda_all}
\resizebox{0.95\textwidth}{!}{%
\begin{tabular}{@{}llccc@{}}
\toprule
\textbf{Sparsity Weight ($\lambda$)} & \textbf{Encoder} & \textbf{Structure (RF Std)} & \textbf{Reconstruction Loss} & \textbf{Sparsity Loss} \\
\midrule
\textbf{$\lambda = 0.0$} & Baseline & 0.049821 & 0.003512 & 0.002510 \\
\textit{(No Constraint)} & Random & 0.042503 & 0.003621 & 0.001905 \\
& Linear & 0.054112 & 0.003498 & 0.002285 \\
& Poisson & 0.041550 & 0.003905 & \textbf{0.000008} \\
& Dynamic Dissipative ($\delta=10.0$) & 0.047015 & 0.003840 & 0.000812 \\
& Dynamic Critical ($\delta=0.0$) & 0.036880 & 0.003865 & 0.001055 \\
& \textbf{Dynamic Transition ($\delta=2.0$)} & \textbf{0.176904} & 0.003710 & 0.000920 \\
& Dynamic Expansive ($\delta=-1.5$) & 0.046233 & \textbf{0.001275} & 1.850210 \\
\midrule
\textbf{$\lambda = 0.1$} & Baseline & 0.048941 & 0.002277 & 0.001549 \\
& Random & 0.042101 & 0.002528 & 0.001410 \\
& Linear & 0.071794 & 0.002119 & 0.001752 \\
& Poisson & 0.043657 & 0.003882 & \textbf{0.000069} \\
& Dynamic Dissipative ($\delta=10.0$) & 0.063901 & 0.002914 & 0.000763 \\
& Dynamic Critical ($\delta=0.0$) & 0.041037 & 0.003009 & 0.001006 \\
& \textbf{Dynamic Transition ($\delta=2.0$)} & \textbf{0.292495} & 0.002719 & 0.001107 \\
& Dynamic Expansive ($\delta=-1.5$) & 0.040490 & \textbf{0.001279} & 0.184287 \\
\midrule
\textbf{$\lambda = 0.5$} & Baseline & 0.043694 & 0.003216 & 0.002427 \\
& Random & 0.035226 & 0.003396 & 0.001973 \\
& Linear & 0.057604 & 0.003072 & 0.002518 \\
& Poisson & 0.046205 & 0.003908 & \textbf{0.000006} \\
& Dynamic Dissipative ($\delta=10.0$) & 0.053488 & 0.003751 & 0.000578 \\
& Dynamic Critical ($\delta=0.0$) & 0.037390 & 0.003769 & 0.000664 \\
& \textbf{Dynamic Transition ($\delta=2.0$)} & \textbf{0.236729} & 0.003429 & 0.001379 \\
& Dynamic Expansive ($\delta=-1.5$) & 0.042407 & \textbf{0.001278} & 0.860399 \\
\midrule
\textbf{$\lambda = 1.0$} & Baseline & 0.041741 & 0.003534 & 0.002529 \\
& Random & 0.034058 & 0.003656 & 0.001896 \\
& Linear & 0.054055 & 0.003506 & 0.002270 \\
& Poisson & 0.049168 & 0.003908 & \textbf{0.000008} \\
& Dynamic Dissipative ($\delta=10.0$) & 0.050199 & 0.003858 & 0.000804 \\
& Dynamic Critical ($\delta=0.0$) & 0.036652 & 0.003872 & 0.001046 \\
& \textbf{Dynamic Transition ($\delta=2.0$)} & \textbf{0.205833} & 0.003722 & 0.000868 \\
& Dynamic Expansive ($\delta=-1.5$) & 0.043464 & \textbf{0.001276} & 1.716117 \\
\midrule
\textbf{$\lambda = 3.0$} & Baseline & 0.039928 & 0.003777 & 0.002443 \\
& Random & 0.034596 & 0.003790 & 0.002142 \\
& Linear & 0.050401 & 0.003767 & 0.002183 \\
& Poisson & 0.044609 & 0.003907 & \textbf{0.000019} \\
& Dynamic Dissipative ($\delta=10.0$) & 0.047220 & 0.003895 & 0.000209 \\
& Dynamic Critical ($\delta=0.0$) & 0.036943 & 0.003923 & 0.001738 \\
& \textbf{Dynamic Transition ($\delta=2.0$)} & \textbf{0.202009} & 0.003870 & 0.000640 \\
& Dynamic Expansive ($\delta=-1.5$) & 0.043081 & \textbf{0.001255} & 5.003095 \\
\bottomrule
\end{tabular}%
}
\end{table*}

\begin{table}[]
\centering
\caption{Robustness validation on the STL-10 dataset ($96 \times 96$ patches) using $\lambda=0.1$. 
\textit{Transition} regime persists superior structural organization on higher-resolution natural images.}
\label{tab:ablation_stl10}
\begin{tabular}{@{}lccc@{}}
\toprule
\textbf{Encoder} & \textbf{Structure (RF Std)} & \textbf{Reconstruction Loss} & \textbf{Sparsity Loss} \\
\midrule
\textbf{Dynamic Transition} & \textbf{0.1892} & 0.0051 & 0.0011 \\
Dynamic Dissipative & 0.0433 & 0.0055 & 0.0010 \\
Baseline & 0.0381 & 0.0049 & 0.0031 \\
Dynamic Expansive & 0.0390 & \textbf{0.0024} & 1.8920 \\
Poisson & 0.0415 & 0.0058 & \textbf{0.0001} \\
\bottomrule
\end{tabular}
\end{table}

\textbf{Fine-Grained Landscape Validation.}
\label{app:landscape}
To rigorously decouple the effects of dynamics from explicit regularization, we conducted a high-resolution parameter sweep over the dynamical control parameter $\delta \in [-1.0, 5.0]$ under varying sparsity weights ($\lambda$). Figure \ref{fig:scan_grid} reveals in-depth interaction patterns between dynamical complexity and learnability.

While conventional methods suggest that explicit sparsity constraints (e.g., L1 regularization) are necessary to induce structured receptive fields, the most significant finding is revealed by the progression from $\lambda=1.0$ (Fig.~\ref{fig:scan_grid}a) to $\lambda=0.0$ (Fig.~\ref{fig:scan_grid}c): the structural quality of the learned features actually improves as the explicit constraint is removed. As shown in Fig.~\ref{fig:scan_grid}, the peak structural score increases from $\sim 0.47$ ($\lambda=1.0$) to $\sim 0.61$ ($\lambda=0.0$). This demonstrates that the \textit{Transition} dynamics ($\delta \approx 2.5$) do not merely "support" feature learning; they act as an advantageous, intrinsic constraint that can be as effective as artificial regularization. This finding provides evidence for the biological plausibility of our model, suggesting that biological systems, operating without explicit global loss functions, can rely on metabolic and physical constraints to shape connectivity. 

Furthermore, we observe a local minimum at the precise center of the critical regime ($\delta \approx 2.6$), characterized by a sharp spike in \textit{Reconstruction Loss} (blue curve) coincident with a dip in \textit{RF Structure} (red curve). This phenomenon likely reflects a trade-off between dynamical complexity and linear decodability: at $\delta \approx 2.6$, the system generates trajectories that are topologically rich but highly non-linear. The linear decoder struggles to reconstruct these complex signals (high loss), temporarily disrupting feature condensation. Consequently, the optimal zones for representation learning naturally settle at the margins of this critical peak ($\delta \approx 2$ and $3$), where the system balances rich dynamical entropy with sufficient regularity for decoding.

\textbf{Spectral Analysis and Scale-Space Invariance.}
\label{app:spectral}
To investigate the physical mechanism underlying the structural emergence, we performed a spectral analysis of the dynamical trajectories generated by the encoder. We estimated the Power Spectral Density (PSD) using Welch's method~\cite{welch2003use} and extracted two key metrics: the \textit{Spectral Centroid}, representing the weighted mean frequency of the signal; and the \textit{Spectral Entropy}, quantifying the complexity of the power distribution.

\begin{figure*}[ht!]
    \centering
    \includegraphics[width=\textwidth]{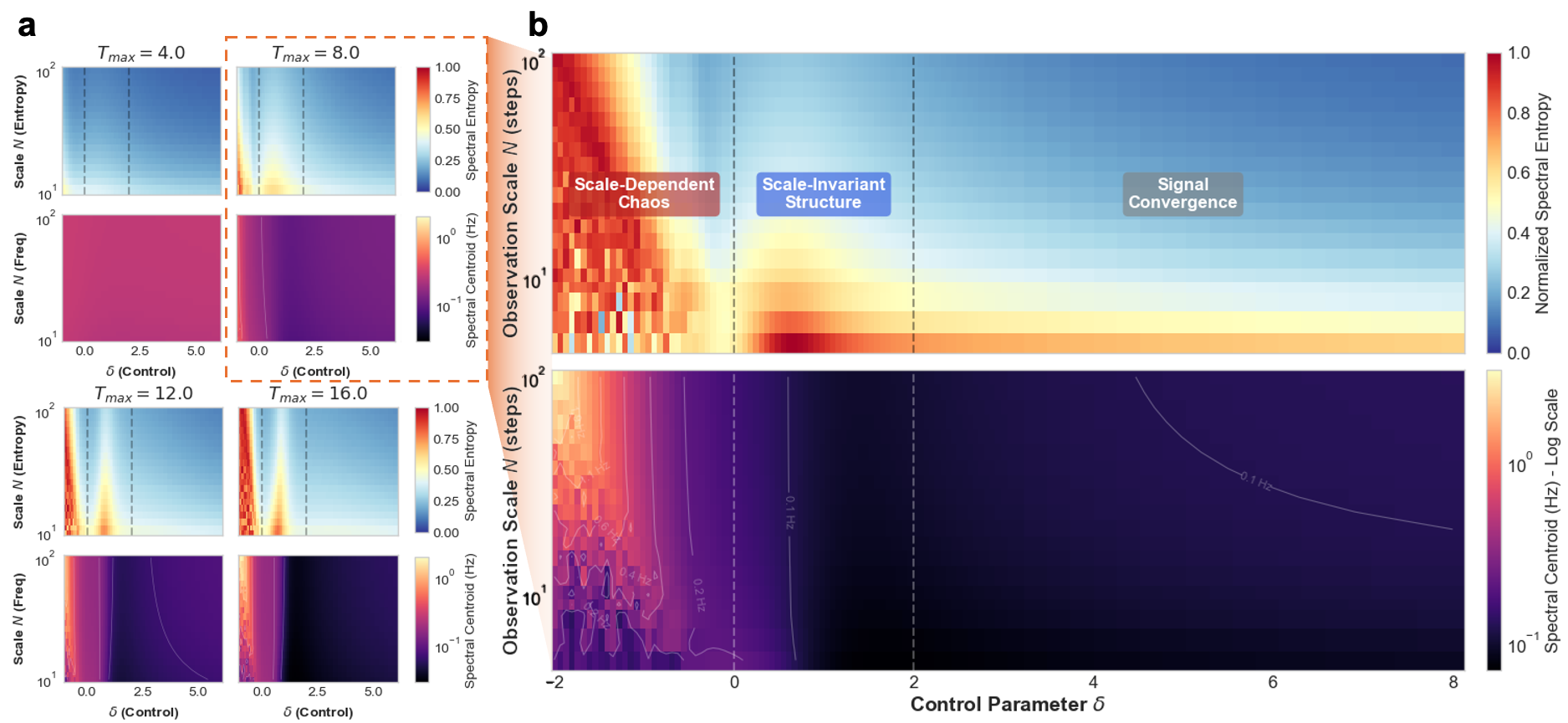} %
    \caption{\textbf{Mechanism of Spectral Alignment and Invariance.} 
    \textbf{(a)} Scale-space heatmaps across varying physical timescales ($T_{max}$) demonstrate the robustness of this mechanism. The dark vertical band in the transition region indicates that the "Low-Frequency, High-Structure" valley is \textit{scale-invariant}, persisting stably across different observation windows. This confirms that the generalization capability arises from locking onto intrinsic invariants rather than transient artifacts.
    \textbf{(b)} Spectral analysis reveals a unique signature in the \textit{Transition} regime ($\delta \approx2$): it minimizes the \textit{Frequency Centroid} (bottom) to align with the network's spectral bias, while simultaneously maintaining high \textit{Spectral Entropy} (top) to preserve structural complexity.}
    \label{fig:spectral_mechanism}
\end{figure*}

\textit{Scale-Space Invariance (Fig.~\ref{fig:spectral_mechanism}A).} To verify that these spectral properties are intrinsic to the dynamical regime rather than artifacts of specific hyperparameters, we performed a multi-scale sweep across observation scales ($N$) and physical evolution times ($T_{max} \in \{4, 8, 12, 16\}$). This analysis reveals a persistent spectral signature localized strictly within the transition regime, characterized by the stable coexistence of low dominant frequencies and high spectral entropy. Furthermore, this signature remains robust across varying timescales, confirming that transition dynamics actively suppress the accumulation of chaotic divergence. The sharp contrast between this ordered state and the high-frequency stochasticity of the expansive regime establishes it as a robust computational spot for generalization capacity.

\textit{Spectral Properties (Fig.~\ref{fig:spectral_mechanism}B).} We analyzed the \textit{Spectral Entropy} and \textit{Dominant Frequency} across the dynamical spectrum. \textit{Expansive} regime ($\delta < 0$) is characterized by maximal entropy and high frequency, resembling broadband white noise, which is difficult for neural networks to regularize. \textit{Dissipative} regime ($\delta \gg 2$) exhibits minimal entropy, indicating information loss. \textit{Transition} regime ($\delta \in [0, 2]$) represents a critical state: it creates a signal that is structurally complex (intermediate-to-high entropy) yet temporally structured (locked to low frequencies).

\subsection{Experiment 3: Behavioral Robustness}
\label{app:exp3}

This appendix provides detailed experimental setups and supplementary analyses for the behavioral robustness experiments described in Section~\ref{sec:robustness}. We present two complementary validation approaches: encoding-level (using Duffing transformation) and architecture-level (varying membrane leak parameter $\beta$).

\subsubsection{Encoding-Level Experiments}
\label{app:exp3_encoding}

\textbf{Setup.} 
Agents were trained exclusively in an "Easy" environment (pole length 0.5m, gravity $9.8~m/s^{2}$) based on the \textit{CartPole-v1} task~\cite{towers2024gymnasium}. They were evaluated zero-shot on three physical properties to create progressively challenging environments while maintaining the same state-space structure, as presented in Table~\ref{tab:cartpole_arch_config}.

\begin{table}[h]
\centering
\caption{\texttt{CartPole} difficulty configurations. Environmental stochasticity increases through pole length, pole mass, force noise, and initial range.}
\label{tab:cartpole_arch_config}
\small
\begin{tabular}{lcccc}
\toprule
\textbf{Difficulty} & \textbf{Pole Length (m)} & \textbf{Pole Mass (kg)} & \textbf{Force Noise} & \textbf{Init Range} \\
\midrule
Easy (training) & 0.5 & 0.1 & 0.0 & 0.05 \\
Medium & 0.8 (+60\%) & 0.3 (+200\%) & 0.005 & 0.08 \\
Hard & 1.2 (+140\%) & 0.5 (+400\%) & 0.01 & 0.10 \\
Very Hard & 1.5 (+200\%) & 0.7 (+600\%) & 0.015 & 0.12 \\
\bottomrule
\end{tabular}
\end{table}

These perturbations alter the system's moment of inertia and responsiveness without changing the observation space or action set, providing a test of learned policies' ability to generalize across physical parameter shifts~\cite{tobin2017domain}.

All policy networks (SNN and MLP) shared an identical fully-connected topology of two hidden layers with 128 units each. This width was chosen to ensure sufficient network capacity, guaranteeing that performance differences would arise from the quality of the input encoding and architectural synergy, rather than from an under-parameterized model bottleneck. We use the REINFORCE algorithm~\cite{williams1992simple} with Adam optimizer ($lr=10^{-3}$, $\gamma=0.99$)~\cite{kingma2014adam}. All results are averaged over 10 independent runs.

For the \textit{Baseline MLP} agent, this resulted in a (4 $\to$ 128 $\to$ 128 $\to$ 2) architecture processing the raw 4-dimensional environment state. For SNN and encoded-MLP agents, each of the four state variables was used to initialize a 3D oscillator mentioned in sections above, creating a 12-dimensional input vector (4 states $\times$ 3 dims), resulting in a (12 $\to$ 128 $\to$ 128 $\to$ 2) topology. The SNN policy network operated with a LIF decay rate of $\beta=0.95$. The MLP models used ReLU activations.

To ensure rigorous comparison, we incorporated \textit{Layer Normalization} within the MLP hidden layers~\cite{ba2016layer}. This design choice was driven by our preliminary ablations: without extrinsic normalization, the stateless MLP suffered rapid numerical divergence when processing the high-variance \textit{Expansive} ($\delta = -1.5$) encoding. Furthermore, while the \textit{Transition} and \textit{Dissipative} regimes achieved comparable mean performance without normalization, they exhibited significantly amplified inter-trial variance. We therefore standardized on Layer Normalization to eliminate optimization instability as a confounder, ensuring that performance differences reflect genuine representational capabilities rather than gradient issues. In contrast, the SNN required no such extrinsic stabilization; its intrinsic leaky-integration mechanism ($\beta=0.95$) provides natural temporal smoothing and dynamic stability. 

Training followed two distinct protocols corresponding to the results in Table~\ref{tab:rl_results}:
\begin{itemize}
    \item \textit{Fixed-Budget (2000 Eps.):} Agents were trained for a strict, fixed budget of 2,000 episodes ($\mathtt{early\_stopping=False}$). This protocol ensures all agents are compared using identical training resources, isolating the impact of encoding and architecture on generalization efficiency.
    \item \textit{Sufficient Training:} Agents were trained for a maximum of 5,000 episodes with an evaluation-based early stopping mechanism ($\mathtt{early\_stopping=True}$). During training, the agent's policy was evaluated every 20 episodes. A policy was considered 'solved' and training was stopped if its mean evaluation reward met or exceeded 195.0 for five consecutive evaluation intervals. This protocol compares the final, converged generalization capability of each agent.
\end{itemize}

\subsubsection{Architecture-Level Experiment Setup}
\label{app:exp3_arch}

To validate that the constraint principle generalizes beyond specific dynamical encoding, we conducted a $\beta$-sweeps where agents received raw state histories without any external dynamical transformation. This isolates the architectural contribution of the SNN's intrinsic temporal dynamics.

\textbf{Setup.} We tested on two tasks of increasing complexity. For \texttt{CartPole}, we used the same difficulty progression as the encoding-level experiments (Table~\ref{tab:cartpole_arch_config}). For \texttt{LunarLander-v3}, we introduced environmental stochasticity through wind, turbulence, and modified landing constraints (Table~\ref{tab:lunarlander_config}).

\begin{table}[h]
\centering
\caption{\texttt{LunarLander} difficulty configurations. Environmental stochasticity increases through wind forces, turbulence, initial velocity perturbations, and narrower landing zones.}
\label{tab:lunarlander_config}
\small
\begin{tabular}{lccccc}
\toprule
\textbf{Difficulty} & \textbf{Gravity Scale} & \textbf{Wind Power} & \textbf{Turbulence} & \textbf{Init Vel Scale} & \textbf{Landing Zone} \\
\midrule
Easy (training) & 0.8 & 0.0 & 0.0 & 0.3 & 2.0$\times$ \\
Medium & 0.9 & 3.0 & 0.3 & 0.6 & 1.5$\times$ \\
Hard & 1.0 & 6.0 & 0.7 & 0.9 & 1.2$\times$ \\
Very Hard & 1.1 & 8.0 & 1.0 & 1.2 & 0.9$\times$ \\
\bottomrule
\end{tabular}
\end{table}

All networks shared a two-layer topology with 256 hidden units. For baseline comparisons, we implemented:
\begin{itemize}
    \item \textit{MLP}: Feedforward network with LayerNorm and ReLU activations, processing single-step observations.
    \item \textit{LSTM}: LSTM layers with LayerNorm, processing sequential observations with hidden state persistence across timesteps.
\end{itemize}

For SNN architectures, we implemented two variants to disentangle the effects of membrane dynamics and recurrent connectivity:
\begin{itemize}
    \item \textit{Leaky SNN}: Feedforward LIF network where temporal integration occurs solely through membrane potential dynamics. The membrane update follows $\text{mem}_{t+1} = \beta \cdot \text{mem}_t + W \cdot x_t$, with surrogate gradient (fast sigmoid, slope=25) for backpropagation through spikes.
    \item \textit{RLeaky SNN}: Recurrent LIF network with additional lateral connections within each layer. The membrane update becomes $\text{mem}_{t+1} = \beta \cdot \text{mem}_t + W_{\text{ff}} \cdot x_t + W_{\text{rec}} \cdot s_t$, where $s_t$ denotes the spike output. This provides an additional memory pathway beyond membrane potential.
\end{itemize}

Weight initialization followed orthogonal initialization with gain $\sqrt{2}$ for hidden layers and gain 0.01 for output layers. For RLeaky networks, recurrent weights were initialized with spectral radius 0.9 to ensure stable dynamics.

The choice of RL algorithm was driven by task complexity. For \texttt{CartPole} (4D state, 200-step episodes), vanilla REINFORCE~\cite{williams1992simple} provided sufficient learning signal. For \texttt{LunarLander} (8D state, 1000-step episodes), the higher-dimensional state space and longer episodes required PPO~\cite{schulman2017proximal} with Generalized Advantage Estimation (GAE, $\lambda=0.95$) for stable training. Critically, neither algorithm introduces confounding temporal mechanisms: both remain policy gradient methods without learned world models, ensuring that performance differences reflect architectural properties rather than algorithmic artifacts.

Table~\ref{tab:training_config} summarizes the hyperparameters for each task. SNN agents were allocated more training episodes to account for slower initial learning, while maintaining identical evaluation protocols.

\begin{table}[h]
\centering
\caption{Training configurations for architecture-level experiments.}
\label{tab:training_config}
\small
\begin{tabular}{lcc}
\toprule
\textbf{Parameter} & \textbf{CartPole (REINFORCE)} & \textbf{LunarLander (PPO)} \\
\midrule
Learning rate (actor) & $5 \times 10^{-4}$ & $3 \times 10^{-4}$ \\
Learning rate (critic) & --- & $1 \times 10^{-3}$ \\
Gradient clipping & 1.0 & 0.5 \\
Hidden dimension & 256 & 256 \\
Max episodes (MLP/LSTM) & 3,000 & 15,000 \\
Max episodes (SNN) & 6,000 & 15,000 \\
Convergence threshold & 195.0 & 200.0 \\
Batch size & --- & 2048 \\
PPO epochs & --- & 10 \\
Clip epsilon & --- & 0.2 \\
\bottomrule
\end{tabular}
\end{table}

After training, each agent was evaluated for 100 episodes on each difficulty level without exploration noise. We report mean reward, standard deviation, and success rate (fraction of episodes achieving reward $\geq$ 195 for CartPole, $\geq$ 200 for LunarLander). The generalization gap is defined as $\text{Gap} = \text{Reward}_{\text{Easy}} - \text{Reward}_{\text{Very Hard}}$.

\textbf{Constraint versus Memory.}
\label{app:rleaky}
A central question in interpreting our results is whether the observed generalization benefits stem from \textit{temporal constraint} (phase space contraction through dissipation) or simply from \textit{temporal memory} (information retention capacity). This distinction is critical: if generalization arises from memory accumulation, then architectures with greater memory capacity should exhibit monotonically improved robustness as retention increases. Conversely, if constraint drives generalization, we expect a non-monotonic relationship where excessive information retention (high $\beta$) permits overfitting to transient features.

RLeaky SNNs provide a critical test case to disentangle these mechanisms. Unlike feedforward Leaky SNNs, which rely solely on membrane potential for temporal integration, RLeaky SNNs incorporate lateral recurrent connections that provide an additional memory pathway through persistent hidden states. This architecture increases the network's capacity to retain information across timesteps~\cite{maass2002real}.

\begin{figure}[ht!]
    \centering
    \includegraphics[width=0.9\textwidth]{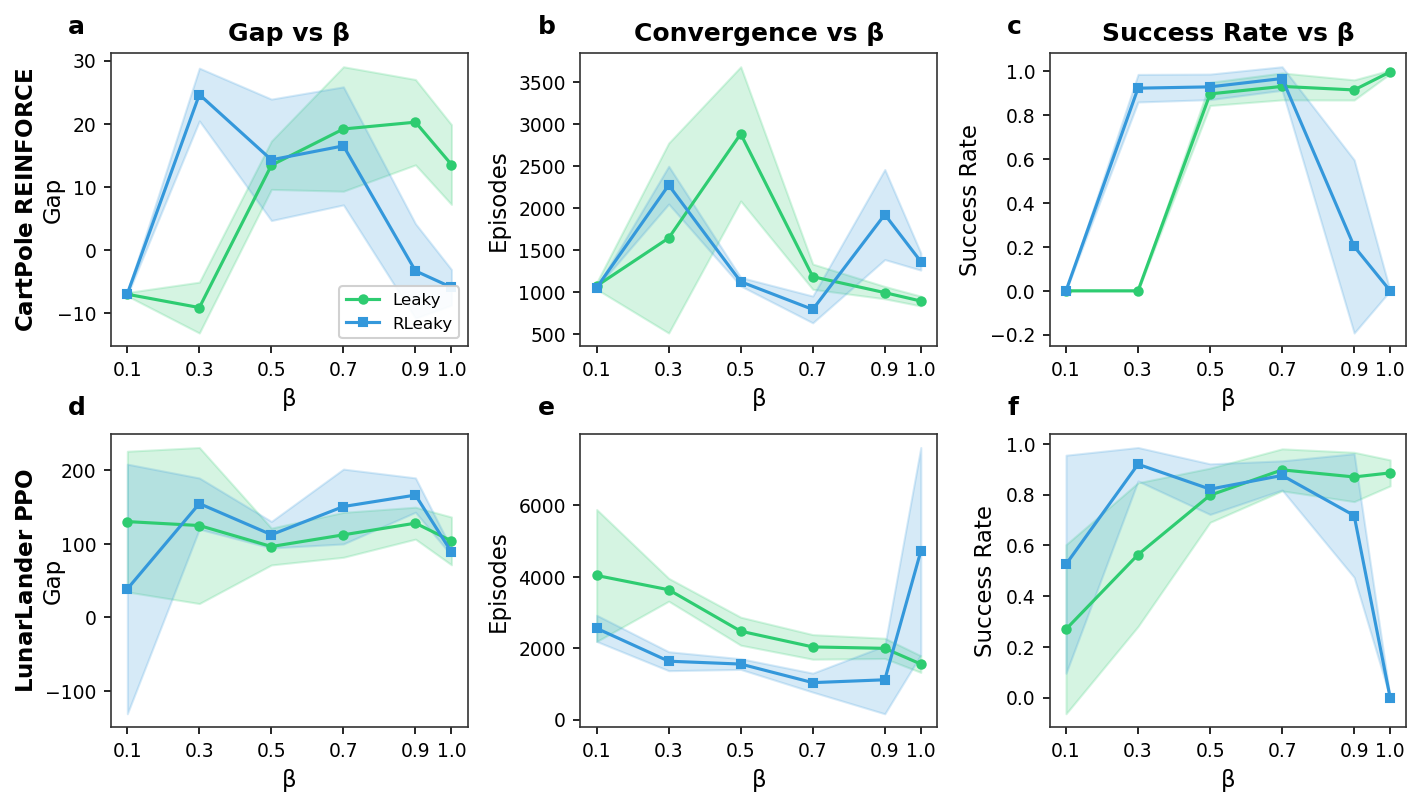}
    \caption{\textbf{$\beta$-sweep comparison between Leaky and RLeaky SNNs.} Top row: \texttt{CartPole} (REINFORCE). Bottom row: \texttt{LunarLander} (PPO). \textbf{(a, d)} Generalization gap versus $\beta$. Both architectures reveal non-monotonic optima. Note that the apparent low gap for RLeaky at $\beta=1.0$ on LunarLander (d) is an artifact of training failure (see f). \textbf{(b, e)} Episodes to convergence. Higher $\beta$ generally accelerates learning, but extreme values trigger instability. \textbf{(c, f)} Success rate on Easy environment. Leaky SNNs are robust across $\beta \ge 0.5$, whereas RLeaky SNNs exhibit significant instability at both extremes, indicating that recurrent memory pathways require stronger dissipative constraints to remain stable.}
    \label{fig:beta_sweep_full}
\end{figure}

\begin{figure}[ht!]
    \centering
    \includegraphics[width=0.9\textwidth]{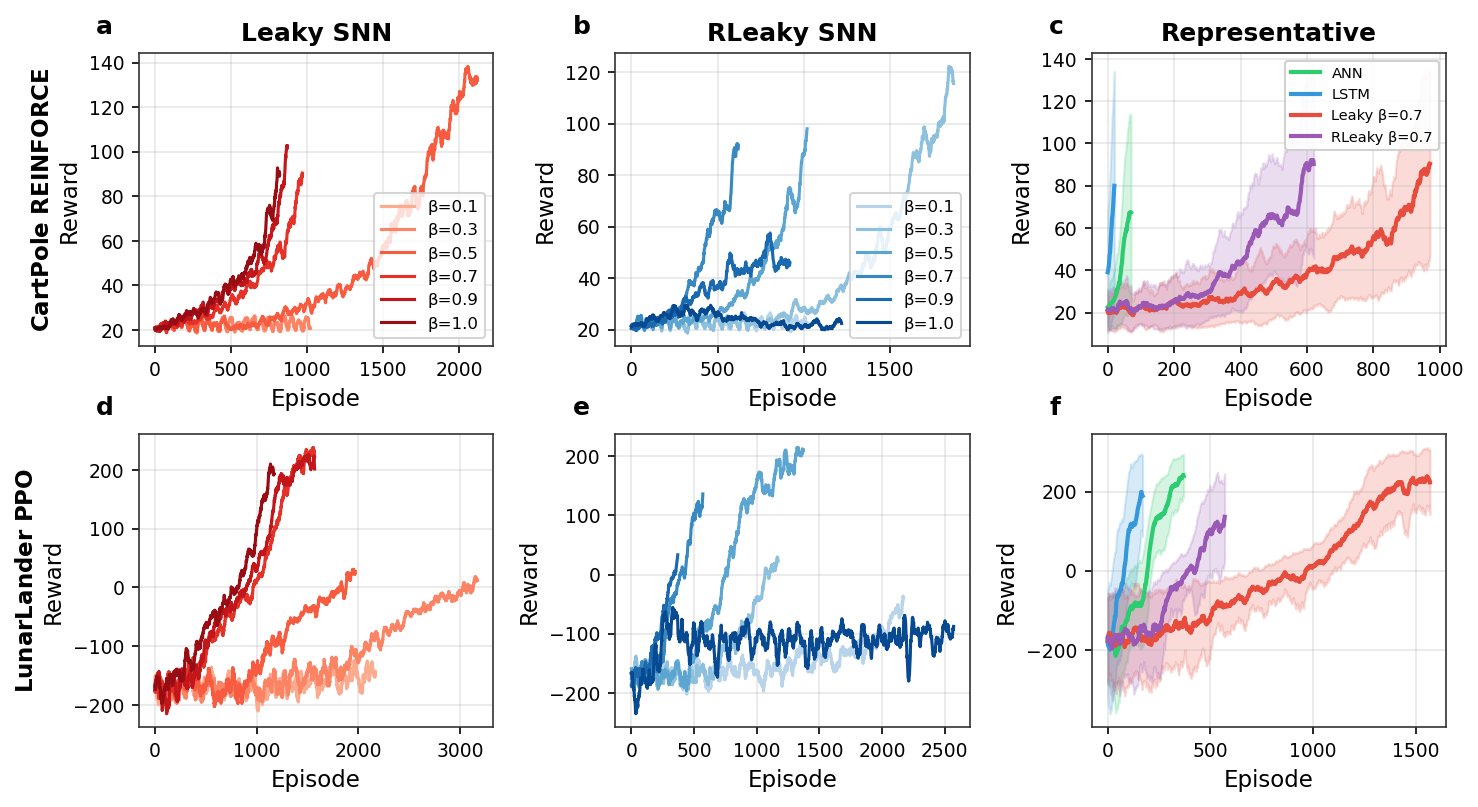}
    \caption{\textbf{Training dynamics across architectures and tasks.} Top row (a--c): \texttt{CartPole} with REINFORCE. Bottom row (d--f): \texttt{LunarLander} with PPO. \textbf{(a, d)} Leaky SNN training curves sorted by $\beta$: stronger retention (higher $\beta$, darker colors) consistently accelerates convergence speed. \textbf{(b, e)} RLeaky SNN training curves exhibit higher variance and instability, particularly at extreme $\beta$ values on LunarLander. \textbf{(c, f)} Representative model comparison with confidence intervals. Baselines (ANN, LSTM) converge rapidly but generalize poorly (shortcut learning). In contrast, the robust Leaky $\beta=0.5$ agent (red) shows a slower, more gradual learning curve, empirically illustrating the cost of abstraction: slow learning associated with robustness.}
    \label{fig:training_curves}
\end{figure}

Table~\ref{tab:rleaky_full} presents complete $\beta$-sweep results for RLeaky SNNs across both tasks, with ANN and LSTM baselines for reference. Figures~\ref{fig:beta_sweep_full} and~\ref{fig:training_curves} visualize the relationship between $\beta$, generalization gap, learning dynamics, and training stability.

\begin{table}[h]
\centering
\caption{RLeaky SNN results. Recurrent variants exhibit higher sensitivity to hyperparameter extremes compared to feedforward SNNs. Low or High $\beta$ values ($\leq 0.3$ or $\ge 0.9$) lead to training instability (failed convergence).}
\label{tab:rleaky_full}
\small
\begin{tabular}{llcccc}
\toprule
\textbf{Task} & \textbf{Architecture} & \textbf{Easy Reward} & \textbf{V. Hard Reward} & \textbf{Gap} $\downarrow$ & \textbf{Conv. Eps.} \\
\midrule
\multirow{8}{*}{\textbf{CartPole}} 
 & \textit{ANN (baseline)} & $199.0 \pm 2.0$ & $60.3 \pm 50.4$ & 138.7 & 100 \\
 & \textit{LSTM (baseline)} & $169.0 \pm 60.5$ & $127.0 \pm 65.2$ & 42.0 & 50 \\
 & RLeaky SNN ($\beta=0.1$) & $8.3 \pm 0.1^\dagger$ & $15.3 \pm 0.2$ & --- & 1050 \\
 & RLeaky SNN ($\beta=0.3$) & $196.3 \pm 2.7$ & $171.7 \pm 6.3$ & 24.6 & 2250 \\
 & \textbf{RLeaky SNN ($\beta=0.5$)} & $198.1 \pm 1.5$ & $\mathbf{183.8 \pm 9.7}$ & $\mathbf{14.3}$ & 1100 \\
 & RLeaky SNN ($\beta=0.7$) & $199.4 \pm 1.0$ & $182.9 \pm 8.9$ & 16.5 & 750 \\
 & RLeaky SNN ($\beta=0.9$) & $65.5 \pm 71.1^\dagger$ & $68.8 \pm 63.7$ & --- & 2050 \\
 & RLeaky SNN ($\beta=1.0$) & $23.3 \pm 19.1^\dagger$ & $29.2 \pm 16.4$ & --- & 1350 \\
\midrule
\multirow{8}{*}{\textbf{LunarLander}} 
 & \textit{ANN (baseline)} & $238.9 \pm 11.2$ & $58.6 \pm 40.5$ & 180.3 & 400 \\
 & \textit{LSTM (baseline)} & $220.0 \pm 14.7$ & $26.9 \pm 27.8$ & 193.1 & 200 \\
 & RLeaky SNN ($\beta=0.1$) & $21.7 \pm 292.2^\dagger$ & $-17.1 \pm 145.4$ & --- & 2400 \\
 & RLeaky SNN ($\beta=0.3$) & $241.0 \pm 18.6$ & $86.0 \pm 38.7$ & 155.0 & 1600 \\
 & \textbf{RLeaky SNN ($\beta=0.5$)} & $222.8 \pm 27.2$ & $\mathbf{110.2 \pm 33.7}$ & $\mathbf{112.6}$ & 1600 \\
 & RLeaky SNN ($\beta=0.7$) & $240.6 \pm 14.1$ & $89.6 \pm 47.3$ & 151.0 & 1000 \\
 & RLeaky SNN ($\beta=0.9$) & $179.6 \pm 95.5^*$ & $13.1 \pm 80.1$ & 166.5 & 800 \\
 & RLeaky SNN ($\beta=1.0$) & $-80.5 \pm 6.8^\dagger$ & $-169.5 \pm 11.0$ & --- & 3200 \\
\bottomrule
\multicolumn{6}{l}{\footnotesize $^\dagger$Training failed to converge. Gap not meaningful. $^*$High variance/Partial convergence.}
\end{tabular}
\end{table}

The comparison between Leaky and RLeaky reveals several important patterns:

\begin{itemize}
    \item \textit{Sensitivity and Stability Window.} While both architectures achieve their respective optimal generalization at a similar characteristic timescale ($\beta \approx 0.5$), they differ fundamentally in stability. Feedforward Leaky SNNs maintain robust performance across a broader range of regimes ($\beta \in [0.5, 1.0]$), achieving a superior generalization gap of \textbf{96.5} on \texttt{LunarLander}. In contrast, RLeaky SNNs exhibit a significantly narrower stability window: they require strict dissipative constraints to function, and their performance degrades rapidly outside the optimum (Gap: 112.6 at $\beta=0.5$, deteriorating to training failure at $\beta=1.0$). This suggests that the additional memory introduced by recurrent connections acts as an expansive force; to prevent chaotic divergence, the network must rely heavily on the dissipative $\beta$ constraint to maintain a stable computational regime.
    
    \item \textit{The Cost of Generalization.} Figure~\ref{fig:training_curves}\textbf{(c, f)} reveals a stark contrast in the temporal structure of learning. Baseline models (ANN, LSTM) converge rapidly (within 400 episodes) but plateau at suboptimal generalization levels, symptomatic of "shortcut learning" that fits transient statistics. In contrast, robust SNN agents (particularly at $\beta=0.5-0.7$) exhibit a much slower, gradual learning curve (requiring 1000+ episodes). This empirically illustrates the "cost of abstraction": the extraction of invariant features is inherently more computationally demanding than the memorization of surface statistics.
    
    \item \textit{Inefficacy of Unconstrained Memory.} RLeaky SNNs possess greater theoretical memory capacity than Leaky SNNs, yet they achieve \textit{worse} generalization (Gap 112.6 vs. 96.5 on \texttt{LunarLander}, see Table~\ref{tab:rleaky_full}) and higher variance (wider confidence intervals in Figure~\ref{fig:training_curves}). This indicates that simply adding retention pathways does not confer robustness. Instead, excess memory without sufficient constraint increases optimization difficulty, leading to erratic learning trajectories.
\end{itemize}

These non-monotonic relationships between $\beta$ and generalization results across different architectures support our central claim: \textit{constraint induces invariance, not memory accumulation.} Invariant representations emerge through gradual consolidation under constraints, whereas unconstrained learning leads to fragile, context-specific solutions. The additional recurrent pathway in RLeaky SNNs expands the hypothesis space without providing additional constraints, thereby increasing optimization difficulty (visible as training instability and variance) without improving generalization. This mechanism operates independently of memory capacity and represents a distinct computational principle from information retention.

\subsection{PAC-Bayesian Analysis of Dynamical Regimes}
\label{app:pacbayes}

\textbf{Setup.} To quantitatively link dynamical regimes to generalization guarantees, we performed a PAC-Bayesian analysis using a Bayesian spiking neural network (Bayesian SNN) trained on the same dataset and setting as in Experiment~1. The dissipation parameter $\delta$ spans the key regimes:
\[
\delta \in \{-1.5, -1.0, 0.0, 1.0, 2.0, 5.0, 10.0\},
\]
covering expansive, transition, and strongly dissipative dynamics (see Table~\ref{table:d1_oscillator_dynamics_transposed}). We train a two-layer Bayesian SNN with a LIF decay rate of $\beta=0.95$. The weight matrices of both layers are equipped with Gaussian variational posteriors,
\[
w \sim q(w) = \mathcal{N}(\mu, \sigma_q^2), \quad \log \sigma_q^2 \text{ learned,}
\]
and are optimized via the reparameterization trick~\cite{blundell2015weight, kingma2013auto}. The network integrates spikes over $5$ simulation steps to produce logits, trained with cross-entropy loss. Training uses the Adam optimizer (lr $=10^{-3}$), batch size $32$, and $100$ epochs per regime.

\textbf{Dynamics-induced prior.}
To connect the input dynamics to the prior in the PAC-Bayesian framework, we define for each regime a zero-mean Gaussian prior
\[
p_\delta(w) = \mathcal{N}\!\big(0,\sigma_p^2(\delta)\big),
\]
where the prior variance is tied to the global Lyapunov sum $\Sigma_\lambda(\delta)$ of the encoder through
\[
\sigma_p^2(\delta) = \sigma_0^2 \exp\!\big(\Sigma_\lambda(\delta)\, T_{\text{enc}}\big),
\]
with baseline variance $\sigma_0^2 = 1.0$. Here, $T_{\text{enc}}=4.0$ denotes the physical evolution time of the dynamical encoding (consistent with Experiment~1). Expansive dynamics ($\delta < 0$) yield $\Sigma_\lambda > 0$ and thus a diffuse prior (large $\sigma_p^2$), whereas dissipative dynamics ($\delta > 0$) yield $\Sigma_\lambda < 0$ and a concentrated prior (small $\sigma_p^2$). For numerical stability, we clip the exponent to $[-5,5]$, which preserves the relative ordering of regimes while avoiding degenerate variances.

\textbf{PAC-Bayesian bound and metrics.}
For each regime, we estimate the empirical training and test errors,
\[
\hat{L}_{\text{train}} = \frac{1}{m_{\text{train}}}\sum\nolimits_i \mathbb{1}[\hat{y}_i \neq y_i], \quad
\hat{L}_{\text{test}} = \frac{1}{m_{\text{test}}}\sum\nolimits_j \mathbb{1}[\hat{y}_j \neq y_j],
\]
where $m_{\text{train}}$ and $m_{\text{test}}$ are the number of training and testing samples, respectively. We define the \emph{generalization gap} as
\[
\text{Gap} = \hat{L}_{\text{test}} - \hat{L}_{\text{train}}.
\]
We further compute the Kullback--Leibler divergence between posterior and prior,
\[
\mathrm{KL}(q\|p_\delta) = \sum_k \mathrm{KL}\Big(\mathcal{N}(\mu_k, \sigma_{q,k}^2)\,\big\|\, \mathcal{N}(0,\sigma_p^2(\delta))\Big),
\]
summing over all network parameters. Given $\hat{L}_{\text{train}}$, training set size $m=m_{\text{train}}$, and confidence parameter $\delta_{\text{conf}}=0.05$, we evaluate the standard PAC-Bayesian upper bound~\cite{mcallester1998some} on the true risk $L(q)$, which holds with probability at least $1-\delta_{\text{conf}}$:
\begin{equation}
L(q) \;\le\; \hat{L}_{\text{train}}
+ \sqrt{\frac{\mathrm{KL}(q\|p_\delta) + \ln\!\big(2\sqrt{m}/\delta_{\text{conf}}\big)}{2(m-1)}}.
\label{eq:pacbayes-bound}
\end{equation}
Finally, to probe the stability of the optimization dynamics, we train a non-Bayesian SNN (same architecture, SGD optimizer) on the same encodings and record the gradient norm $\|\nabla \mathcal{L}\|$ per minibatch across $200$ epochs. We summarize the gradient statistics by the mean $\mu_{\text{grad}}$ and the coefficient of variation
\[
\mathrm{CV}_{\text{grad}} = \frac{\sigma_{\text{grad}}}{\mu_{\text{grad}} + \varepsilon},
\]
which quantifies the relative variability of gradients across the training trajectory, where $\varepsilon=10^{-8}$ is a small constant for numerical stability.

\textbf{Results.}
Table~\ref{tab:pacbayes_metrics} summarizes the key quantities across dynamical regimes. As dissipation increases, the KL divergence exhibits a sharp reduction from the expansive to the transition regime, before stabilizing in the dissipative region:

\begin{table}[h]
\centering
\caption{PAC-Bayesian analysis revealing the stability-performance trade-off. While Expansive dynamics ($\delta < 0$) achieve lower empirical error (Test Err), they incur a massive complexity penalty (KL), resulting in a loose generalization bound (PAC Bound). The Transition regime ($\delta=2.0$, bolded) minimizes the PAC Bound, representing the optimal theoretical guarantee despite the local performance valley.}
\label{tab:pacbayes_metrics}
\begin{tabular}{lccccc}
\toprule
$\delta$ & $\Sigma_\lambda$ & KL Divergence & Test Err & Gen. Gap & \textbf{PAC Bound} $\downarrow$ \\
\midrule
$-1.5$ & $+3.0$  & $4.6\times 10^4$ & 0.041 & 0.010 & 4.333 \\
$-1.0$ & $+2.0$  & $4.6\times 10^4$ & 0.032 & 0.007 & 4.324 \\
$0.0$  & $0.0$   & $1.5\times 10^4$ & 0.035 & 0.014 & 2.449 \\
$1.0$  & $-2.0$  & $2.3\times 10^3$ & 0.083 & 0.012 & 1.040 \\
$\mathbf{2.0}$ & $\mathbf{-4.0}$ & $\mathbf{2.3\times 10^3}$ & \textbf{0.069} & \textbf{0.008} & \textbf{1.019} \\
$5.0$  & $-10.0$ & $2.3\times 10^3$ & 0.067 & 0.001 & 1.035 \\
$10.0$ & $-20.0$ & $2.4\times 10^3$ & 0.054 & -0.005 & 1.039 \\
\bottomrule
\end{tabular}
\end{table}

Spearman correlation between $\delta$ and KL is negative ($\rho = -0.679$), qualitatively confirming that stronger dissipation yields more concentrated posteriors relative to the dynamics-induced prior. Across all regimes, the PAC-Bayesian bound~\eqref{eq:pacbayes-bound} remains a valid upper bound on the empirical gap (bound validity rate $=100\%$ in our runs), although numerically loose, as is typical in high-dimensional settings.

The gradient analysis, presented in Table~\ref{tab:gradient_stats}, reveals a complementary picture. The coefficient of variation of gradient norms follows a non-monotonic, U-shaped profile across regimes:

\begin{table}[h]
\centering
\caption{Optimization stability analysis via gradient statistics. The transition regime ($\delta = 2.0$) achieves the lowest coefficient of variation ($\mathrm{CV}_{\text{grad}}$), indicating the most stable training dynamics compared to the high variability in expansive regimes and the noise dominance in strongly dissipative regimes.}
\label{tab:gradient_stats}
\begin{tabular}{lccc}
\toprule
$\delta$ & Regime & $\mu_{\text{grad}}$ & $\mathrm{CV}_{\text{grad}}$ \\
\midrule
$-1.5$ & Expansive   & 0.558 & 2.492 \\
$-1.0$ & Expansive   & 0.537 & 1.957 \\
$0.0$  & Critical    & 0.554 & 1.093 \\
$1.0$  & Transition  & 0.721 & 0.916 \\
$2.0$  & Transition  & 0.732 & \textbf{0.822} \\
$10.0$ & Dissipative & 0.655 & 1.132 \\
\bottomrule
\end{tabular}
\end{table}

The transition regime around $\delta = 2.0$ presents the lowest $\mathrm{CV}_{\text{grad}}$ and the highest mean gradient magnitude $\mu_{grad}$, indicating a critical point that balances signal preservation and noise suppression. Expansive regimes exhibit large, highly variable gradients, consistent with the chaotic amplification of perturbations, whereas strongly dissipative regimes suppress gradients to the point that noise dominates, slightly increasing relative variability again.

\textbf{Interpretation.}
Taken together, these results provide a PAC-Bayesian interpretation of the ``performance valley'' observed in the main experiments. At the representational and behavioral levels, networks driven by transition dynamics do not maximize instantaneous in-distribution accuracy; both expansive ($\delta < 0$) and strongly dissipative ($\delta \gg 0$) regimes can achieve comparable or even higher training performance in specific tasks. However, the PAC-Bayesian analysis shows that:
\begin{enumerate}
    \item KL$(q\|p_\delta)$ decreases sharply as we move from expansive to dissipative dynamics, indicating that dissipative temporal structure acts as an implicit prior that constrains the posterior toward a smaller, more stable region of parameter space.
    \item Within the dissipative side, the transition regime minimizes gradient variability and achieves small generalization gaps while retaining sufficient model flexibility. It sits at a ``sweet spot'' where the network is neither driven into chaotic weight excursions (expansive) nor frozen into an over-contracted state (strongly dissipative).
\end{enumerate}
In this sense, the transition regime is a \emph{generalization optimum} rather than a performance optimum: it is the point where the trade-off between fit (training error) and stability (small KL and low $\mathrm{CV}_{\text{grad}}$) is most favorable. From the perspective of dynamical alignment, the edges of the spectrum ($\delta \ll 0$ and $\delta \gg 0$) support specialized computation: rapid discrimination or robust energy-efficient stabilization, while the transition regime provides the structural conditions under which learned solutions generalize across dynamical contexts. PAC-Bayesian theory thus formalizes the intuitive statement that ``constraints breed generalization'' by showing that the temporal constraints induced by dissipative dynamics manifest as tighter priors, more stable training dynamics, and theoretically justified generalization guarantees.

\printbibliography[title={Appendix References},resetnumbers]
\end{refsection} 
\end{document}